\documentclass[a4paper]{cas-sc}
\usepackage{fancyhdr}
\usepackage[numbers]{natbib}
\usepackage{hyperref}
\hypersetup{colorlinks=true, citecolor=blue, linkcolor=blue, urlcolor=blue}
\usepackage{tabularx}

\usepackage{diagbox}
\usepackage{amsthm}
\usepackage{etoolbox}
\usepackage{float}
\usepackage{enumitem}
\floatstyle{ruled}
\newfloat{algorithm}{tbp}{loa}
\floatname{algorithm}{Algorithm}
\usepackage{ragged2e}
\newcolumntype{Y}{>{\centering\arraybackslash}p{3.5cm}}
\newcolumntype{Z}{>{\centering\arraybackslash}X}

\makeatletter
\renewenvironment{proof}[1][\proofname]{\par
  \pushQED{\qed}
  \normalfont \topsep6\p@\@plus6\p@\relax
  \trivlist
  \item[\hskip\labelsep \textit{#1}:]\ignorespaces
}{
  \popQED\endtrivlist\@endpefalse
}
\makeatother

\newtheoremstyle{boldremark}
  {}{}
  {\normalfont}
  {}
  {\bfseries}
  {.}
  {0.5em}
  {}

\theoremstyle{boldremark}
\newtheorem{remark}{Remark}

\newtheorem{theorem}{Theorem}
\usepackage{amsmath,amssymb,amsthm}

\theoremstyle{definition}
\newtheorem{definition}{Definition}

\theoremstyle{plain}

\newtheorem{proposition}{Proposition}

\theoremstyle{plain}
\newtheorem{assumption}{Assumption}

\usepackage{fancyhdr}

\shorttitle{CARGO}
\shortauthors{Teng et al.}

\usepackage{amsmath,amsfonts,amssymb}
\usepackage{bm}
\usepackage{booktabs}
\usepackage{multirow}
\usepackage{makecell}
\ExplSyntaxOn
\cs_set:Npn \__first_footerline: {\small\sffamily Teng~et~al.:~Preprint}
\ExplSyntaxOff
\begin{document}

\let\WriteBookmarks\relax

\let\printorcid\relax
\raggedbottom

\title[mode = title]{CARGO: Causal Attribution and Responsibility-Guided Preventive Safe Multi-Agent Reinforcement Learning for Cooperative Driving}

\author[1]{Jingjia Teng}
\author[1]{Yang Li}
\cormark[1]
\author[1]{Wenjie Huang}

\cortext[1]{Corresponding author.}

\address[1]{State Key Laboratory of Advanced Design and Manufacturing Technology for Vehicle, College of Mechanical and Vehicle Engineering, Hunan University, Changsha 410082, China.}

\begin{abstract}
This study presents Causal Attribution and Responsibility-Guided Optimization (CARGO), a multi-agent reinforcement learning framework for cooperative on-ramp merging in mixed traffic. A shared latent-context posterior and an action-conditioned response model estimate directed behavioral influence, with posterior uncertainty propagated into interaction-edge selection. Complementary attribution mechanisms couple signed influence with marginal team utility for cooperative reward shaping and assign responsibility from positive excess tail risk relative to a safe reference policy. Conditional diffusion generates future responses for action-dependent risk estimation. Responsibility-guided preventive optimization calibrates agent-specific risk-reduction requirements to local capacity within a policy trust region, using reserved safety margins, adaptive requirement scaling, and joint feasibility verification. Theoretical analysis establishes risk monotonicity and sufficient local conditions for non-decreasing environmental return under explicit assumptions. Experiments in highway-env across three traffic densities, together with controlled graph-recovery and attribution studies, support the effectiveness of the framework. Under high-density traffic, CARGO increases the success rate by 66.04\% and reduces the crash rate by 56.23\% relative to the best evaluated baseline for each metric.

\end{abstract}


\maketitle
\hypersetup{pdfauthor={Jingjia Teng, Yang Li, Wenjie Huang},pdfsubject={Research preprint}}

\section{Introduction}
\subsection{Background and Related Work}
Multi-agent reinforcement learning (MARL) supports cooperative driving by learning coordinated vehicle behavior through interaction \cite{Chendong2023TITS,GU2023AI}. In mixed-traffic on-ramp merging, connected and automated vehicles (CAVs) must coordinate with human-driven vehicles (HDVs) under partial observability. A vehicle may decelerate to yield or because of shared downstream congestion, while the vehicle exposed to collision risk need not be its source. Reliable learning therefore requires distinguishing behavioral influence from shared responses, attributing cooperative gains and avoidable risks, and translating attribution into jointly feasible policy corrections.

\textbf{Interaction modeling.}
Directed prediction models and learned coordination graphs represent vehicle dependencies and support cooperative decisions \cite{sun2022m2i,rowe2023fjmp,duan2025inferring,zhang2025influence}. Counterfactual action comparisons and conditional mutual information quantify behavioral influence \cite{jaques2019social,du2024situation}. Latent-variable inference addresses unobserved factors \cite{jiang2024discovering}, while multi-head counterfactual estimation supports uncertainty-aware communication graphs \cite{cg2ac}. These approaches motivate accounting for both shared context and the reliability of inferred interactions. CARGO propagates uncertainty in a history-inferred shared latent context into directed influence estimates and selects edges using a posterior confidence criterion.

\textbf{Contribution and responsibility.}
Counterfactual credit assignment evaluates individual actions against alternative actions or model-generated trajectories \cite{foerster2018counterfactual,chai2024aligning}. Agent-specific effects and effect decomposition further analyze causal propagation \cite{triantafyllou2024agent,triantafyllou2025counterfactual}. Influence-based reward shaping can credit an agent for teammates' contributions \cite{gonzalez2025influence}, whereas counterfactual responsibility identifies sources of safety violations or evaluates alternative driving motions \cite{niu2024responsible,hsu2023responsibility}. Influence strength alone does not determine team benefit, and incurred cost does not identify avoidable risk caused by an action. CARGO couples signed behavioral influence with marginal team utility and separately attributes positive excess tail risk relative to a safe reference policy.

\textbf{Safe policy optimization.}
Shielding and action projection modify behavior before execution \cite{alshiekh2018shielding,xiao2023shielding,kochdumper2023action}, while constrained policy optimization regulates learning updates \cite{achiam2017constrained,GU2023AI,zhang2024scalable}. Safety-aware trust-region geometry and finite-horizon constraint estimation support preventive updates \cite{milosevic2025embedding,dai2024safe}; proactive penalties provide another constraint-handling mechanism \cite{yang2026pcpo}. Responsibility-aware controllers allocate collision avoidance or behavioral adjustments \cite{lyu2023risk,cosner2023learning,remy2025learning}, and SafeDiSTL decomposes shared violation costs using temporal-logic robustness and action sensitivities \cite{zheng2026safedistl}. However, attribution alone does not ensure that differentiated corrections are jointly achievable. CARGO calibrates responsibility requirements to local reduction capacity, then verifies them together with all agents' surrogate safety constraints.

\subsection{Contributions}
\begin{figure}[pos=!tb]
\centering
\includegraphics[width=1\linewidth]{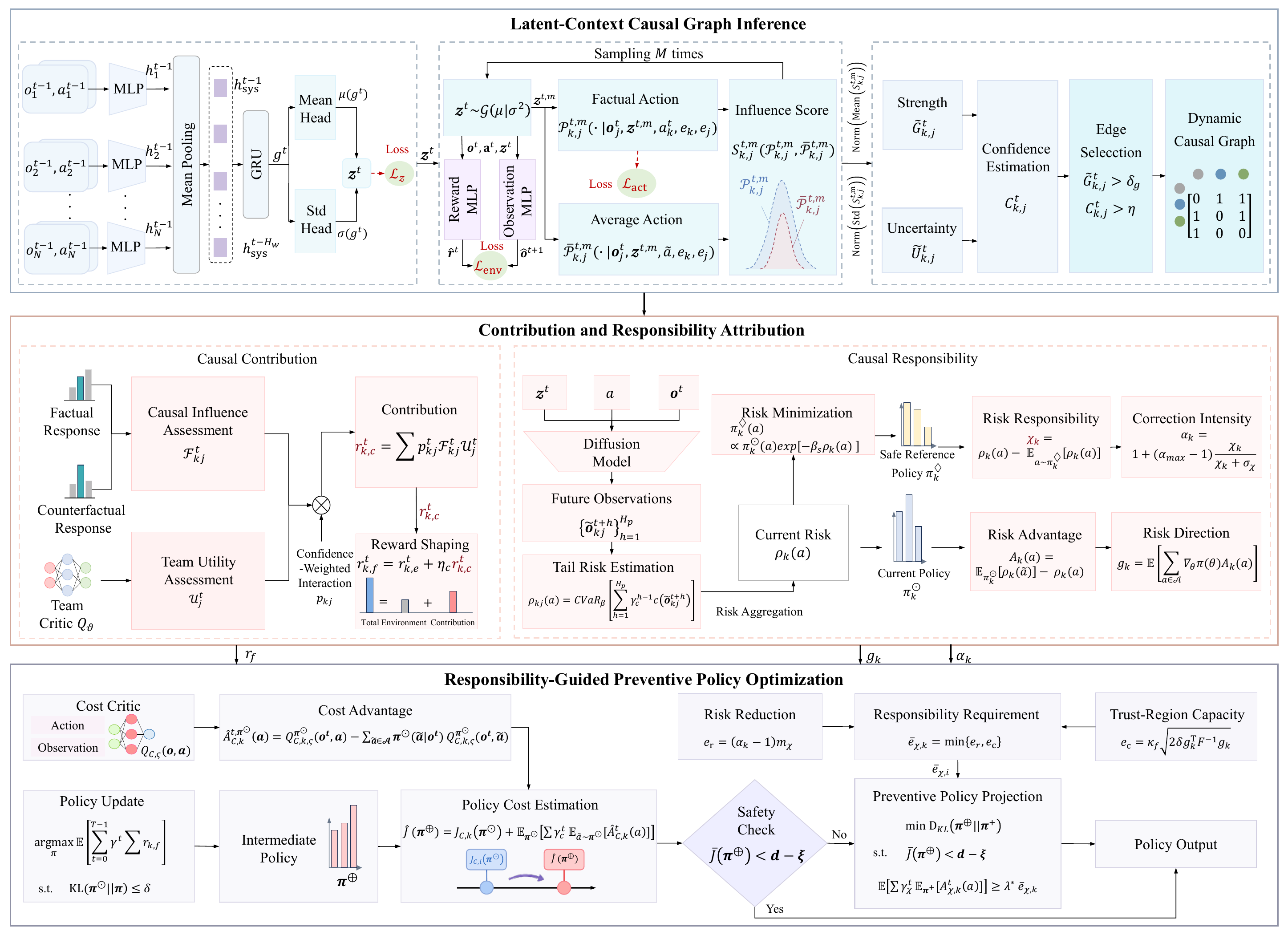}

\caption{Overview of CARGO. Interaction histories are encoded
into a shared latent posterior. Multiple posterior samples
support uncertainty-aware influence estimation and dynamic
causal graph construction. Causal contribution combines behavioral
influence with team utility to shape cooperative rewards,
while causal responsibility quantifies avoidable tail risk
and determines policy correction intensity and direction.
An intermediate policy is accepted if its estimated costs
preserve the reserved safety margins; otherwise, preventive
projection enforces safety constraints and feasible
responsibility-based risk-reduction requirements.}
\label{fig:Framework}

\end{figure}

As illustrated in Fig.~\ref{fig:Framework}, CARGO connects interaction modeling, complementary attribution, and preventive optimization. Its contributions are:
\begin{itemize}
\item An uncertainty-aware interaction model that propagates a shared latent-context posterior into influence estimates and confidence-based edge selection.
\item Complementary attribution of cooperative contribution and safety responsibility: signed influence is weighted by marginal team utility, while diffusion-generated responses estimate positive excess tail risk relative to a safe reference policy.
\item Responsibility-guided preventive optimization that calibrates agent-specific risk-reduction requirements to trust-region capacity and coordinates corrections through adaptive scaling and joint feasibility verification.
\end{itemize}
The analysis establishes risk monotonicity and sufficient local conditions for environmental-return preservation. Evaluation focuses on highway-env merging across three traffic densities, with targeted tests of graph recovery, contribution incentives, responsibility attribution, and policy correction.

\section{Problem Formulation and Overall Framework}\label{sec:framework}
\subsection{Problem Formulation}
We model the heterogeneous on-ramp merging task as a Constrained Decentralized Partially Observable Markov Decision Process (\textbf{Constrained Dec-POMDP}), defined by the tuple
$\langle \mathcal{N}, \mathcal{S}, \mathcal{A}, P, \Omega, \mathcal{O}, R, \mathcal{C}, \gamma_r,\gamma_c \rangle$.
Let $\mathcal{V}$ denote the set of all vehicles, including both CAVs and HDVs, and let $\mathcal{N}$ denote the set of controllable CAV agents.
The global state space is defined as the Cartesian product of individual vehicle state spaces, i.e., $\mathcal{S}=\prod_{i\in\mathcal{V}}\mathcal{S}_i$,
while the joint action space is defined only over CAV agents as $\mathcal{A}=\prod_{i\in\mathcal{N}}\mathcal{A}_i$.
HDVs are not modeled as learning agents. Instead, they are treated as non-controllable environmental participants whose states are included in the global state and whose motions affect the transition dynamics.
The transition probability distribution is denoted by
$P(\mathbf{s}'\mid\mathbf{s},\mathbf{a})$, where $\mathbf{a}=(a_1,\ldots,a_N)\in\mathcal{A}$
denotes the joint action of all CAV agents.
Moreover, $\Omega=\prod_{i\in\mathcal{N}}\Omega_i$ denotes the joint observation space. The set of local observation functions is defined as $\mathcal{O}=\{O_i\}_{i\in\mathcal{N}}$, where $O_i:\mathcal{S}\times\mathcal{A}\to\Delta(\Omega_i)$ denotes the local observation probability distribution of agent $i$.
The reward function $R$ is constructed from agent-wise rewards, and the team reward is given by the sum of all CAV rewards. $\mathcal{C}=\{C_i\}_{i\in\mathcal{N}}$ denotes the set of agent-wise cost functions, and $\gamma_r,\gamma_c\in[0,1)$ denote the reward and cost discount factors, respectively.

\textbf{State Space, Action Space and Observation Space}.
The state of vehicle $i\in\mathcal{V}$ is denoted as $\mathbf{s}_i^{t}= [x_i^{t},y_i^{t},v_i^{x,t},v_i^{y,t}]^\top$,
where $x_i^{t}$ and $y_i^{t}$ denote the longitudinal and lateral positions, respectively, and $v_i^{x,t}$ and $v_i^{y,t}$ denote the corresponding longitudinal and lateral velocities. Each CAV agent $i$ is assumed to perceive only a localized traffic region. The neighboring vehicle set $\mathcal{M}_i^{t}$ contains up to four nearest vehicles within a 150~m longitudinal range \cite{Chaoyu2020TITS}. Their observations are ordered by increasing distance from the ego vehicle, with unused slots zero-padded when fewer than four neighbors are available, yielding a fixed-dimensional observation vector.
The observation $\boldsymbol{o}_i^{t}$ of CAV agent $i$ consists of its own state $\mathbf{s}_i^{t}$ and the relative observations $\boldsymbol{o}_{i,j}^{t}$ of its neighboring vehicles and is expressed as
$\boldsymbol{o}_i^{t} = \big[\mathbf{s}_i^{t}, \{\boldsymbol{o}_{i,j}^{t}\}_{j \in \mathcal{M}_i^{t}}\big]$.
For neighboring vehicle $j$, the relative observation is defined as
$\boldsymbol{o}_{i,j}^{t} = \big[\Delta x_{i,j}^{t}, \Delta y_{i,j}^{t}, \Delta v_{i,j}^{x,t}, \Delta v_{i,j}^{y,t}\big]^\top$.
Each CAV agent $i$ selects an action $a_i\in\mathcal{A}_i$ from the discrete action space defined as
$\mathcal{A}_i=\{a^{\leftarrow},a^{\rightarrow},a^{\uparrow},a^{\downarrow},a^{\oslash}\}$,
where $a^{\leftarrow}$, $a^{\rightarrow}$, $a^{\uparrow}$, $a^{\downarrow}$, and $a^{\oslash}$ denote lane change left, lane change right, accelerating, decelerating, and maintaining speed, respectively.

\textbf{Reward Function}. It is designed to promote safe and efficient cooperative merging through four weighted components that encourage collision avoidance, speed maintenance, appropriate headway, and timely merging. Specifically, the environment reward $r_{i,e}^t$ of CAV agent $i$ at time step $t$ is defined as:
\begin{equation}
\begin{split}
r_{i,e}^t=w_c r_i^{c,t}  + w_v r_i^{v,t} + w_h r_i^{h,t} + w_m r_i^{m,t},
\end{split}
\label{eq:reward}
\end{equation}
where $r_i^{c,t}$, $r_i^{v,t}$, $r_i^{h,t}$, and $r_i^{m,t}$ denote the collision penalty, velocity reward, time-headway reward, and late-merging penalty, respectively, and $w_c$, $w_v$, $w_h$, and $w_m$ are the corresponding positive weights.  Specifically, $r_i^{c,t}=-1$ if a collision occurs and $0$ otherwise.

The velocity reward is defined as $r_i^{v,t}=(v_i^{t}-v_{\min})/(v_{\max}-v_{\min})$, where $v_i^{t} \in [v_{\min},v_{\max}]$.

The time-headway reward is defined as $r_i^{h,t}= \log\left( \frac{d_{\mathrm{head},i}^{t}+\epsilon_h}{t_{\text{THW}} v_i^{t}+\epsilon_h} \right)$, where $d_{\mathrm{head},i}^{t}$ denotes the distance headway of agent $i$. $t_{\text{THW}}$ is a constant denoting the time-headway threshold. $\epsilon_h$ is a small positive constant for numerical stability. The time-headway reward is set to zero when no front vehicle
is available.

To encourage ramp vehicles to merge into the mainline early, we define the late-merging penalty as $r_i^{m,t}=-\exp[-(\ell_i^{t}-L_{\text{ramp}})^2/\sigma_L^2]$, where
$\ell_i^{t}=\max\{x_i^{t}-x_L,0\}$ denotes the traveled distance along the ramp, $x_L$ denotes the ramp entrance position, $L_{\text{ramp}}$ is the ramp length, and $\sigma_L$ is a positive scaling parameter. The late-merging penalty is applied only to ramp vehicles before merging completion.

\textbf{Cost Function}.
It is designed to quantify safety violations and potential collision risks during vehicle interactions. It combines a collision cost with a time-to-collision (TTC)-based interaction cost that considers both front and rear neighboring vehicles. Specifically, the cost of CAV agent $i$ at time step $t$ is defined as:
\begin{equation}
c_i^t=w_{\mathrm{col}} c_i^{{\mathrm{col}},t}+w_{\mathrm{TTC}} c_i^{{\mathrm{TTC}},t},
\end{equation}
where $w_{\mathrm{col}}$ and $w_{\mathrm{TTC}}$ are the weighting coefficients for the collision cost and the TTC-based interaction cost, respectively. The collision cost $c_i^{{\mathrm{col}},t}$ equals $1$ if a collision occurs and $0$ otherwise. Considering the TTC risk with respect to both the nearest front and rear neighboring vehicles, the TTC-based cost $c_i^{{\mathrm{TTC}},t}$ is given by:
\begin{equation}
\begin{aligned}
c_i^{{\mathrm{TTC}},t}
=&
\sum_{q\in\{f,r\}}
w_q
\mathbb{I}\!\left(\Delta v_{i,j_q}^{q,t}>0\right)
\times
\left[
\min\left(
\frac{\max \{d_{\mathrm{head},i,j_q}^{q,t},d_\epsilon\}}
{\max\{\Delta v_{i,j_q}^{q,t},\kappa_v\}},
T_m
\right)
\right]^{-1},
\end{aligned}
\label{eq:ttc_cost}
\end{equation}
where $q\in\{f,r\}$ indexes the front and rear directions with an available neighboring vehicle; directions without one contribute zero cost, and $j_q$ denotes the nearest neighboring vehicle of agent $i$ in direction $q$. $d_{\mathrm{head},i,j_q}^{q,t}$ is the corresponding longitudinal headway between ego vehicle $i$ and vehicle $j_q$, and $\Delta v_{i,j_q}^{q,t}$ indicates the approaching relative speed.
Specifically, $\Delta v_{i,j_f}^{f,t}=v_i^{x,t}-v_{j_f}^{x,t}$ for the front direction, where $v_{j_f}^{x,t}$ denotes the speed of the front neighboring vehicle $j_f$. $\Delta v_{i,j_r}^{r,t}=v_{j_r}^{x,t}-v_i^{x,t}$ for the rear direction, where $v_{j_r}^{x,t}$ denotes the speed of the rear neighboring vehicle $j_r$. $\Delta v_{i,j_q}^{q,t}>0$ indicates that agent $i$ and the neighboring vehicle $j_q$ in direction $q$ are approaching each other. The function $\mathbb{I}(\cdot)$ is the indicator function, $\kappa_v>0$ and $d_\epsilon>0$ are small constants used to avoid division by zero, $T_m$ is the TTC clipping threshold, and $w_q$ is the directional risk weight.

\textbf{Optimization Objective}.
The objective is to learn decentralized policies for CAV agents that maximize the expected discounted team return. Meanwhile, the expected discounted cost of each agent must remain within the safety limit. The resulting constrained policy optimization problem is formulated as:
\begin{equation}
\begin{aligned}
\max_{\boldsymbol{\pi}}\quad
&
\mathbb{E}_{\boldsymbol{\pi}}
\left[
\sum_{t=0}^{T-1}\gamma_r^t
\sum_{i=1}^Nr_{i,e}^t
\right]
\\
\text{s.t.}\quad
&
\mathbb{E}_{\boldsymbol{\pi}}
\left[
\sum_{t=0}^{T-1}\gamma_c^tc_i^t
\right]
\leq d_i, \quad i\in\mathcal{N},
\end{aligned}
\label{eq:constrained_policy_objective}
\end{equation}
where $\boldsymbol{\pi}=(\pi_1,\ldots,\pi_N)$ denotes the joint policy. Under conditionally independent action selection, the joint action distribution is factorized as $\boldsymbol{\pi}(\mathbf{a}^t\mid\boldsymbol{o}^t)=\prod_{i=1}^{N}\pi_i(a_i^t\mid\boldsymbol{o}_i^t)$, where $\mathbf{a}^t=(a_1^t,\ldots,a_N^t)$ and $\boldsymbol{o}^t=(\boldsymbol{o}_1^t,\ldots,\boldsymbol{o}_N^t)$. Each decentralized policy selects actions based on its local observation, and $T$ denotes the episode horizon. $r_{i,e}^t$ and $c_i^t$ denote the environmental reward and safety cost of agent $i$ at time step $t$, respectively. The environmental rewards define the original cooperative objective and characterize task performance. $d_i$ is the upper limit on the expected discounted cumulative safety cost of agent $i$.

\subsection{Overall Framework}
Fig.~\ref{fig:Framework} connects three training modules: shared-context inference selects reliable interaction edges; contribution and responsibility attribution provide cooperative incentives and risk-correction signals; preventive optimization verifies policy updates against reserved safety margins and capacity-calibrated responsibility requirements. An intermediate cooperative policy is accepted if it satisfies the surrogate constraints; otherwise, projection seeks a verified correction. Deployment uses decentralized policies without the auxiliary causal and risk models.

\section{Latent-Context Causal Graph Inference}
\label{sec:causal_graph}
Shared environmental factors can induce correlated behavior without direct influence. We infer shared context from interaction histories and compare action-dependent responses across posterior samples; influence strength and posterior confidence then determine the retained edges.

\subsection{Uncertainty-Aware Causal Graph Inference}

\textbf{Latent Context Representation.}
Let $\mathcal H_t^-=\{(\mathbf o^\tau,\mathbf a^\tau)\}_{\tau=t-H_w}^{t-1}$ contain joint observations and actions before the current decision, excluding the current source action. A shared MLP encodes each observation--action pair; mean pooling across the $N$ agents gives $h_{\mathrm{sys}}^\tau$, and a GRU summarizes this sequence as $g^t$:
\begin{equation}
h_{\mathrm{sys}}^\tau=\frac1N\sum_{i=1}^N\operatorname{MLP}(\operatorname{Concat}(o_i^\tau,a_i^\tau)),
\qquad g^t=\operatorname{GRU}(h_{\mathrm{sys}}^{t-H_w},\ldots,h_{\mathrm{sys}}^{t-1}).
\label{eq:system_representation}
\end{equation}
The encoder parameterized by $\psi$ defines a diagonal-Gaussian posterior over the $l_z$-dimensional shared context:
\begin{equation}
q_\psi(\boldsymbol z^t\mid\mathcal H_t^-)
=\mathcal G(\boldsymbol z^t;\mu(g^t),\operatorname{diag}(\sigma^2(g^t))),
\qquad \boldsymbol z^{t,m}\sim q_\psi(\cdot\mid\mathcal H_t^-),\quad m=1,\ldots,M.
\label{eq:latent_posterior}
\end{equation}
Here $\mu$ and $\sigma$ are the posterior mean and standard-deviation heads. Multiple samples represent uncertainty about context not uniquely determined by the finite history.

\textbf {Influence Score Computation}.
To quantify directed behavioral influence, we compare the target agent's predicted next-action distribution under the source agent's factual action with an action-averaged reference. This reference averages the predicted responses over all candidate source actions, avoiding reliance on an arbitrarily selected counterfactual action. The resulting distributional deviation is measured by the KL divergence to obtain an action-specific influence score. Its causal interpretation assumes conditional independence between the target agent's potential next-action response and the source action given the target observation and latent context, together with consistency and sufficient action support. Under these assumptions, the estimated influence reflects intervention-induced behavioral changes.

To account for uncertainty in the inferred latent context, we evaluate this influence across posterior samples $\boldsymbol{z}^{t,m}$ drawn from $q_\psi(\cdot\mid\mathcal H_t^-)$. The target observation $o_j^t$ and the same latent sample $\boldsymbol{z}^{t,m}$ are held fixed across all source-action evaluations; the latent context is not re-estimated when the source action is replaced. To distinguish different source--target agent pairs, we incorporate agent identity embeddings into the conditional response model. Specifically, the distribution conditioned on the factual action is defined as:
\begin{equation}
\mathcal{P}_{k,j}^{t,m}(\cdot\mid o_j^t,\boldsymbol{z}^{t,m},a_k^t,e_k,e_j)
=
\operatorname{Softmax}
\left(
\operatorname{MLP}
\left(
\operatorname{Concat}(o_j^t,\boldsymbol{z}^{t,m},a_k^t,e_k,e_j)
\right)
\right),
\label{eq:factual_action_distribution}
\end{equation}
where $\mathcal{P}_{k,j}^{t,m}$ denotes the predicted
next-action distribution of agent $j$ conditioned on
the factual action of agent $k$, and
$\operatorname{Softmax}(\cdot)$ converts the network outputs
into probabilities over the target action space $\mathcal{A}_j$. $e_k$ and $e_j$ denote the learnable identity embeddings of the source and target agents, respectively.
To construct the reference distribution, we replace $a_k^t$
with each possible source action and uniformly average
the resulting predictions while keeping the remaining conditioning
inputs fixed:
\begin{equation}
\overline{\mathcal{P}}_{k,j}^{t,m}(\cdot\mid o_j^t,\boldsymbol{z}^{t,m},e_k,e_j)
=
\frac{1}{|\mathcal{A}_k|}
\sum_{\tilde a\in\mathcal{A}_k}
\operatorname{Softmax}
\left(
\operatorname{MLP}
\left(
\operatorname{Concat}(o_j^t,\boldsymbol{z}^{t,m},\tilde a,e_k,e_j)
\right)
\right),
\label{eq:averaged_action_distribution}
\end{equation}
where $\overline{\mathcal{P}}_{k,j}^{t,m}$ denotes the
action-averaged reference distribution. $\mathcal{A}_k$ is the action space of agent $k$ and $\tilde a\in\mathcal{A}_k$ denotes a candidate action of agent $k$
used to construct the action-averaged reference distribution. We then quantify the influence for each latent sample
using the Kullback--Leibler (KL) divergence:
\begin{equation}
S_{k,j}^{t,m}
=
D_{\mathrm{KL}}
\left[
\mathcal{P}_{k,j}^{t,m}
(\cdot\mid o_j^t,\boldsymbol{z}^{t,m},a_k^t)
\,\middle\|\,
\overline{\mathcal{P}}_{k,j}^{t,m}
(\cdot\mid o_j^t,\boldsymbol{z}^{t,m})
\right],
\label{eq:inter_agent_edge_score}
\end{equation}
where $S_{k,j}^{t,m}$ is the influence score from $k$
to $j$ under the $m$-th latent sample and
$D_{\mathrm{KL}}(\cdot)$ denotes the KL divergence.

\textbf{Posterior-Uncertainty-Aware Graph Construction}.
To construct a dynamic graph that accounts for uncertainty in the inferred environmental context, we aggregate the influence scores across latent posterior samples to estimate both influence strength and variability. These statistics are subsequently used to evaluate posterior confidence and determine the retained interaction edges. A single latent sample cannot
capture how the estimated influence varies across plausible
contexts. We therefore draw $M$ samples from the latent
posterior and compute the edge score for each sample. The influence strength from agent $k$ to agent $j$ is
estimated as the sample mean of the resulting influence scores:
\begin{equation}
G_{k,j}^{t}
=
\frac{1}{M}
\sum_{m=1}^{M} S_{k,j}^{t,m},
\label{eq:causal_strength_mean}
\end{equation}
where $G_{k,j}^{t}$ denotes the estimated influence strength averaged over the $M$ latent posterior samples. The corresponding uncertainty is quantified by the sample standard deviation:
\begin{equation}
U_{k,j}^{t}
=
\sqrt{
\frac{1}{M-1}
\sum_{m=1}^{M}
\left(S_{k,j}^{t,m}-G_{k,j}^{t}\right)^2
},
\label{eq:causal_strength_uncertainty}
\end{equation}
where $U_{k,j}^{t}$ characterizes the variability of the estimated influence across latent posterior samples, reflecting its sensitivity to uncertainty in the inferred environmental context. To assess the relative importance of each interaction within the current traffic context, we normalize both the influence strength and its uncertainty by the mean influence strength across all ordered pairs of distinct agents:
\begin{equation}
\begin{aligned}
\widetilde{G}_{k,j}^{t}
=
\frac{G_{k,j}^{t}}
{\max\left\{
\frac{1}{N(N-1)}
\sum_{u=1}^{N}
\sum_{\substack{v=1\\v\neq u}}^{N}
G_{u,v}^{t},
\epsilon_G
\right\}},
\widetilde{U}_{k,j}^{t}
=
\frac{U_{k,j}^{t}}
{\max\left\{
\frac{1}{N(N-1)}
\sum_{u=1}^{N}
\sum_{\substack{v=1\\v\neq u}}^{N}
G_{u,v}^{t},
\epsilon_G
\right\}},
\end{aligned}
\label{eq:normalized_causal_statistics}
\end{equation}
where $\widetilde{G}_{k,j}^{t}$ and $\widetilde{U}_{k,j}^{t}$
are the normalized influence strength and uncertainty,
respectively. $\epsilon_G>0$ is a small constant for numerical stability. Under a Gaussian approximation of the normalized influence-score distribution
induced by the latent posterior, we estimate the edge confidence as the probability that the normalized influence score exceeds a reference threshold:
\begin{equation}
C_{k,j}^{t}
=
\Phi\left(
\frac{\widetilde{G}_{k,j}^{t}-\delta_c}
{\max\{\widetilde{U}_{k,j}^{t},\epsilon_C\}}
\right),
\label{eq:edge_confidence}
\end{equation}
where $C_{k,j}^{t}$ denotes the Gaussian-approximated posterior exceedance probability.
$\Phi(\cdot)$ is the standard Gaussian cumulative distribution
function. $\delta_c$ is the reference threshold and $\epsilon_C>0$ is a small constant for numerical
stability. Finally, we retain an edge only when both its normalized strength
and confidence exceed the corresponding thresholds:
\begin{equation}
A_{k,j}^{t}
=
\begin{cases}
1,
& \widetilde{G}_{k,j}^{t}>\delta_g,\;
C_{k,j}^{t}>\eta,\\
0, & \text{otherwise},
\end{cases}
\label{eq:binary_graph}
\end{equation}
where $A_{k,j}^{t}$ is the binary adjacency indicator
for the directed edge from agent $k$ to agent $j$
at time step $t$. $\delta_g$ denotes the relative influence strength threshold, and $\eta\in(0,1)$ is the confidence threshold. Under the adopted settings
$\delta_g=\delta_c=0.5$ and $\eta=0.8$,
the confidence criterion implies the strength
criterion. Consequently, edge retention is
effectively determined by the posterior confidence,
which jointly reflects the estimated relative
influence strength and its uncertainty.

\subsection{Latent-Context Learning Objectives}
The latent encoder and prediction networks are jointly trained using
\begin{equation}
\mathcal L_{\mathrm{total}}=\mathcal L_{\mathrm{env}}+\mathcal L_z+\mathcal L_{\mathrm{act}}.
\label{eq:lcgi_total_loss}
\end{equation}
Separate MLPs predict the next joint observation $\widehat{\boldsymbol o}^{t+1}$ and current reward vector $\widehat{\mathbf r}^t$ from $(\boldsymbol o^t,\mathbf a^t,\boldsymbol z^t)$. For a batch of $B$ samples,
\begin{equation}
\mathcal L_{\mathrm{env}}=\frac1{BN}\sum_{b=1}^B
\left[\|\widehat{\boldsymbol o}_b^{t+1}-\boldsymbol o_b^{t+1}\|^2+
\lambda_r\|\widehat{\mathbf r}_b^t-\mathbf r_b^t\|^2\right],
\quad
\mathcal L_z=\frac1B\sum_{b=1}^B D_{\mathrm{KL}}\!\left(q_\psi(\cdot\mid\mathcal H_{t,b}^-)\Vert\mathcal G(0,I)\right).
\label{eq:environment_loss}
\end{equation}
Here $\lambda_r$ weights reward prediction. The conditional response model is supervised by observed next actions:
\begin{equation}
\mathcal L_{\mathrm{act}}=-\frac1{BN(N-1)}\sum_{b=1}^B\sum_{k=1}^N\sum_{j\ne k}
\log\mathcal P_{k,j}^t(a_{j,b}^{t+1}\mid o_{j,b}^t,\boldsymbol z_b^t,a_{k,b}^t,e_{k,b},e_{j,b}).
\label{eq:response_model_loss}
\end{equation}
Latent samples use reparameterization so both prediction objectives update the posterior encoder; $\mathcal L_z$ regularizes it toward a standard Gaussian.

\section{Contribution and Responsibility Attribution}
\label{sec:optimization}
In this section, we exploit the identified causal relationships to determine which agents
contribute to cooperative gains or induce avoidable safety risks based on the inferred dynamic causal graph in~Section \ref{sec:causal_graph}. To this end, we formulate causal contribution and causal
responsibility to guide cooperative incentives and policy updates, respectively.

\begin{definition}[Causal Contribution]
Causal contribution characterizes the assistive role of agent $k$ in shaping team outcomes through its causal influence on the actions of another agent $j$, where $j\neq k$. Agent $k$ makes a positive contribution if its action facilitates team-beneficial actions by agent $j$ or suppresses team-harmful actions, and should therefore be rewarded. Conversely, agent $k$ makes a negative contribution if its action hinders team-beneficial actions or facilitates team-harmful actions, and should therefore be penalized.
\end{definition}

\begin{definition}[Causal Responsibility]
Causal responsibility characterizes the extent to which an agent's actual action exposes itself or other agents to risk that could have been reduced by acting more safely. An agent bears responsibility for this avoidable risk when its actual action induces a higher risk than a safer alternative. The greater the risk that could have been avoided, the greater the agent's causal responsibility.
\end{definition}

\subsection{Causal Contribution-Based Incentive}
To promote cooperation through the inferred causal relationships, we introduce causal contribution to quantify an agent's assistance to the team. This measure combines the agent's influence on others' actions with the marginal team utility of those actions and serves as an agent-specific intrinsic reward. This intrinsic reward is employed as an auxiliary training incentive, whereas the original cooperative task objective remains defined by environmental rewards.

\textbf{Causal Influence Assessment.}
To assess the directed influence of agent $k$ on agent $j$'s next action, we compare the predicted behavioral responses under different source actions while holding the target observation and inferred latent context fixed.
Unlike the nonnegative KL divergence used for
graph construction, a signed log-likelihood ratio
is adopted to distinguish whether the factual source action increases or decreases the predicted probability of the target agent's realized next action. The causal influence of agent $k$ on agent $j$'s next action can be defined as:
\begin{equation}
\mathcal{F}_{kj}^{t}
=
\log
\frac{
\mathcal{P}_{k,j}^{t}
\left(
a_j^{t+1}\mid o_j^t,\boldsymbol z^t,a_k^t,e_k,e_j
\right)
}{
\overline{\mathcal{P}}_{k,j}^{t}
\left(
a_j^{t+1}\mid o_j^t,\boldsymbol z^t,e_k,e_j
\right)
},
\label{eq:causal_influence}
\end{equation}
where $\mathcal{F}_{kj}^{t}$ is the influence score,
$\mathcal{P}_{k,j}^{t}$ is the learned conditional
response distribution, and
$\boldsymbol z^t=\mu(g^t)$ is the posterior mean
of the shared latent context.
The action-averaged reference is defined as
\begin{equation}
\begin{aligned}
&\overline{\mathcal{P}}_{k,j}^{t}
\left(
a_j^{t+1}\mid o_j^t,\boldsymbol z^t,e_k,e_j
\right)=
\frac{1}{|\mathcal A_k|}
\sum_{\widetilde a\in\mathcal A_k}
\mathcal{P}_{k,j}^{t}
\left(
a_j^{t+1}\mid o_j^t,\boldsymbol z^t,
\widetilde a,e_k,e_j
\right),
\end{aligned}
\label{eq:contribution_reference_distribution}
\end{equation}
where $\overline{\mathcal{P}}_{k,j}^{t}$
denotes the action-averaged reference
distribution evaluated at the posterior mean
$\boldsymbol z^t$, and
$\widetilde a\in\mathcal A_k$ is a candidate
action of agent $k$.
A positive $\mathcal{F}_{kj}^{t}$ indicates
that agent $k$'s factual action increases
the model-predicted probability of agent $j$'s
realized next action relative to the
action-averaged reference, whereas a negative
value indicates a decrease in that probability.

\textbf{Team Utility Assessment.}
We evaluate the marginal team utility of agent $j$'s
realized action by comparing its team value with a
policy-averaged baseline while holding the other agents'
actions fixed:
\begin{equation}
\mathcal{U}_j^t
=
Q_{\vartheta,e}\Big(\boldsymbol{o}^{t+1},\left(
a_j^{t+1},
\mathbf{a}_{\mathcal{N}\setminus\{j\}}^{t+1}
\right)\Big)
-
\mathbb{E}_{\tilde a_j^{t+1}\sim\pi_j(\cdot\mid o_j^{t+1})}
\left[
Q_{\vartheta,e}
\left(
\boldsymbol{o}^{t+1},
\left(
\tilde a_j^{t+1},
\mathbf{a}_{\mathcal{N}\setminus\{j\}}^{t+1}
\right)
\right)
\right],
\label{eq:team_utility}
\end{equation}
where $\mathbf{a}_{\mathcal{N}\setminus\{j\}}^{t+1}$ denotes the actions of all agents except agent $j$. The joint action notation
$(a_j^{t+1},
\mathbf{a}_{\mathcal{N}\setminus\{j\}}^{t+1})$
represents the original joint action, while
$(\tilde a_j^{t+1},
\mathbf{a}_{\mathcal{N}\setminus\{j\}}^{t+1})$
denotes the joint action obtained by replacing
only agent $j$'s action with
$\tilde a_j^{t+1}$.
In both cases, the original agent ordering
is preserved. $Q_{\vartheta,e}$ is the centralized team action-value
function computed using environmental rewards and
implemented by a network that maps the joint observation
and joint action to a scalar value through three fully
connected hidden layers with ReLU activation and a
linear output layer. $\vartheta$ denotes the trainable parameters of the centralized team critic.
$\mathcal{U}_j^t$ measures the marginal team utility of $a_j^{t+1}$ for attribution to agent $k$'s action at time $t$. A positive $\mathcal{U}_j^t$ indicates that the realized action is more beneficial to the team than its policy-averaged baseline, whereas a negative value indicates the opposite.

\textbf{Causal Contribution.} Combining the causal facilitation effect and team utility effect, the causal contribution of agent $k$ is defined as:
\begin{equation}
r_{k,c}^{t}
=
\sum_{\substack{j=1,j\neq k}}^{N}
p_{kj}^{t}
\mathcal{F}_{kj}^{t}
\mathcal{U}_{j}^{t},
\label{eq:causal_contribution_reward}
\end{equation}
where $r_{k,c}^{t}$ denotes the causal contribution of agent $k$, and $p_{kj}^{t}=A_{k,j}^{t}C_{k,j}^{t}$ is the confidence-weighted interaction coefficient from agent $k$ to agent $j$. The causal contribution reward is computed only when the next joint action is available. For terminal transitions, the causal contribution reward is set to zero. The shaped reward used for policy training is defined as:
\begin{equation}
r_{k,f}^{t}
=
r_{k,e}^t
+
\eta_cr_{k,c}^{t},
\label{eq:final_reward}
\end{equation}
where $r_{k,f}^{t}$ and $r_{k,e}^t$ represent the shaped training reward and the environment reward, respectively. $\eta_c>0$ is the causal contribution coefficient. For policy training, we incorporate the shaped reward $r_{i,f}^t$ into the reward objective of \eqref{eq:constrained_policy_objective} to construct an auxiliary training surrogate that encourages agents to facilitate beneficial actions and discourage harmful ones. The original constrained task remains defined by the environmental rewards and the original safety-cost limits; reward shaping does not alter these constraints.

\subsection{Causal Responsibility}
To guide safety-oriented policy updates, we define causal responsibility as the avoidable risk of an agent's
factual action relative to the expected risk under a safe reference policy.
Responsibility determines the correction intensity, while risk advantage guides the correction direction, jointly steering policy updates toward lower-risk actions.

\subsubsection{Tail-Risk Estimation}

To assess the risk induced by an agent's action, we use a conditional diffusion model to predict its own future observations and those of other agents under alternative actions. 

\textbf{Diffusion-based Future Response Generation}. For source agent $k$ and recipient agent $j$, the conditioning representation under candidate action
$a\in\mathcal A_k$ is defined as
\begin{equation}
\begin{aligned}
\mathbf c_{kj}^{t}(a)
= {}&\operatorname{MLP}\bigl(\operatorname{Concat}(\,
\mathbf o^{t},\boldsymbol{z}^{t},\bar{\mathbf p}_{k}^{t},\operatorname{OneHot}(a),e_k,e_j\,)\bigr),
\end{aligned}
\end{equation}
where $\mathbf o^t$ is the current joint observation,
and $\boldsymbol{z}^t=\mu(g^t)$ is the posterior mean of the shared
latent context.
$\bar{\mathbf p}_k^t=[\bar p_{k1}^t,\ldots,\bar p_{kN}^t]$
is the interaction-weight vector for source agent $k$,
where $\bar p_{kk}^t=1$ and
$\bar p_{kj}^t=
A_{k,j}^{\tau_g(t)}
C_{k,j}^{\tau_g(t)}$ for $j\neq k$.
Here $\tau_g(t)<t$ denotes the time index
of the most recent completed graph update
available before the current action decision.
Accordingly, the interaction weights are
determined before evaluating the current
candidate actions and remain fixed across
all candidate-action evaluations.
$\operatorname{OneHot}(a)$ encodes the candidate action,
while $e_k$ and $e_j$ are learnable embeddings of the
source and recipient identities, respectively. The conditioning MLP is jointly optimized
with the denoising network through the
diffusion noise-prediction loss.

Conditioned on $\mathbf c_{kj}^{t}(a)$, we generate possible future responses through denoising diffusion
implicit model (DDIM) sampling:
\begin{equation}
\begin{aligned}
\widetilde{\mathbf o}_{kj}^{\,t+1:t+H_p,l}(a)
&=
\operatorname{DDIM}_{\bar\theta_d}
\bigl(\epsilon^l,\mathbf c_{kj}^{t}(a)\bigr)
,\quad
\epsilon^l
\sim\mathcal G(0,I),
\quad l=1,\ldots,L,
\end{aligned}
\label{eq:counterfactual_response}
\end{equation}
where $\widetilde{\mathbf o}_{kj}^{\,t+1:t+H_p,l}(a)$ is the $l$-th predicted observation sequence of agent $j$ under source action $a$, and $H_p$ is the prediction horizon. $L$ denotes the number of generated
future response samples. $\operatorname{DDIM}_{\bar\theta_d}(\cdot)$ denotes
DDIM sampling using the exponential moving average
(EMA) parameters $\bar\theta_d$ of a conditional
denoising network. The denoising network takes the noisy sequence, diffusion-step embedding, and conditioning representation as inputs. The initial Gaussian noise samples are independent across $l$ and shared across candidate actions for the same source--recipient pair. $\mathcal G(0,I)$ denotes the standard multivariate Gaussian distribution, where $I$ is the identity matrix.

\textbf{Conditional Diffusion Training.}
Observed future sequences $\mathbf y_j^t=\mathbf o_j^{t+1:t+H_p}$ are conditioned on factual source actions. At diffusion step $s\in\{1,\ldots,S\}$, set $\mathbf y_{j,s}^t=\sqrt{\bar\alpha_s}\mathbf y_j^t+\sqrt{1-\bar\alpha_s}\varepsilon$, where $\bar\alpha_s$ is the cumulative signal-retention coefficient and $\varepsilon\sim\mathcal G(0,I)$. The masked noise-prediction objective is
\begin{equation}
\mathcal L_{\mathrm{diff}}=\mathbb E_{(t,k,j),s,\varepsilon}
\left[\frac{\|\mathbf M_j^t\odot(\varepsilon-\varepsilon_{\theta_d}(\mathbf y_{j,s}^t,s,\mathbf c_{kj}^t(a_k^t)))\|_2^2}{\|\mathbf M_j^t\|_1+\epsilon_m}\right],
\label{eq:conditional_diffusion_training}
\end{equation}
where $\mathbf M_j^t$ masks missing future observations and $\epsilon_m>0$ stabilizes the denominator. During action comparison, only $a_k^t$ is replaced by the candidate action $a$; all other conditioning inputs remain fixed.

\textbf{Tail-Risk Estimation.}
To evaluate the safety risk of the generated future
responses, we use a separate cost predictor
$c_{\bar\omega}$ to estimate safety
costs from predicted local observations.
During risk evaluation, the cost predictor and
diffusion model use their EMA parameters
$\bar\omega$ and $\bar\theta_d$, respectively.
Risk evaluation is performed only when the complete
prediction horizon of length $H_p$ is valid.
For each generated response, we compute the
discounted cumulative cost:
\begin{equation}
J_{kj}^{t,l}(a)
=
\sum_{h=1}^{H_p}
\gamma_c^{h-1}
\left[
c_{\bar\omega}
\left(
\widetilde{\mathbf o}_{kj}^{\,t+h,l}(a)
\right)
\right]_+,
\label{eq:predicted_response_cost}
\end{equation}
where
$\widetilde{\mathbf o}_{kj}^{\,t+h,l}(a)$
is the predicted observation of recipient agent $j$
at time $t+h$, $\gamma_c\in[0,1)$ is the cost
discount factor, and $[x]_+=\max\{x,0\}$ clips
predicted costs to nonnegative values.
The recipient's tail risk is estimated by averaging
the largest cumulative costs among the generated
responses:
\begin{equation}
\rho_{kj}^{t}(a)
=
\frac{1}{K_\beta}
\sum_{l=1}^{K_\beta}J_{kj}^{t,(l)}(a),
\qquad
K_\beta
=
\max\left\{1,\left\lceil(1-\beta)L\right\rceil\right\},
\label{eq:empirical_counterfactual_cvar}
\end{equation}
where $\rho_{kj}^{t}(a)$ denotes the predicted tail
risk for agent $j$ under source agent $k$'s candidate
action $a$, and $J_{kj}^{t,(l)}(a)$ is the $l$-th
largest cumulative cost among the $L$ generated
responses.
$\beta\in(0,1)$ is the confidence level, and
$K_\beta$ is the number of retained upper-tail
samples.

To evaluate each candidate action from the source
agent's perspective, we combine its own tail risk
with those of other agents, weighted by their
causal relations:
\begin{equation}
\rho_k^t(a)
=
\frac{
\rho_{kk}^t(a)
+
\displaystyle\sum_{j\neq k}
\bar p_{kj}^t\rho_{kj}^t(a)
}{
1+\displaystyle\sum_{j\neq k}\bar p_{kj}^t
},
\label{eq:source_aggregated_tail_risk}
\end{equation}
where $\rho_k^t(a)$ is the aggregated tail-risk
score, $\rho_{kk}^t(a)$ is the source agent's own
tail risk, and $\bar p_{kj}^t\geq0$ is the
interaction weight obtained from the latest
pre-decision graph.
The source agent's own risk is assigned a unit
weight, i.e., $\bar p_{kk}^t=1$.

\textbf{Cost Predictor Training.}
The cost predictor is trained by supervised
regression using post-transition local observations
and their corresponding instantaneous safety costs.
These observations contain the neighboring-vehicle
information required to evaluate the TTC-based
interaction cost.
The training objective is
\begin{equation}
\mathcal L_{\mathrm{cost}}
=
\mathbb E_{
(o_j^{\tau+1},c_j^\tau)\sim\mathcal D
}
\left[
\big(
c_\omega(o_j^{\tau+1})-c_j^\tau
\big)^2
\right],
\label{eq:cost_predictor_training}
\end{equation}
where $\mathcal D$ denotes the collected interaction
data, $o_j^{\tau+1}$ is agent $j$'s local observation
after the transition from time $\tau$ to $\tau+1$,
and $c_j^\tau$ is the corresponding instantaneous
safety cost returned by the environment.
The predictor $c_\omega$, parameterized by $\omega$,
comprises two ReLU-activated hidden layers followed
by a scalar linear output layer.
Its EMA parameters $\bar\omega$ are used for
the tail-risk estimation described above.

\subsubsection{Responsibility Attribution}
To quantify the risk that could have been avoided by changing
the factual action, we construct a safe reference policy
that favors lower-risk actions while remaining close to the
current behavior policy. For $\beta_s>0$, this policy is
obtained by solving the following KL-regularized risk
minimization problem:
\begin{equation}
\pi_k^{\diamondsuit,t}(\cdot\mid o_k^t;\beta_s)
=
\arg\min_{\pi_k^t\in\Delta(\mathcal{A}_k)}
\left\{
\mathbb{E}_{a\sim\pi_k^t(\cdot\mid o_k^t)}
\left[\rho_k^t(a)\right]
+
\frac{1}{\beta_s}
D_{\mathrm{KL}}
\left(
\pi_k^t(\cdot\mid o_k^t)
\,\middle\|\,
\pi_k^\odot(\cdot\mid o_k^t)
\right)
\right\},
\label{eq:unified_safe_reference}
\end{equation}
where $\pi_k^{\diamondsuit,t}$ and $\pi_k^\odot$ denote the
safe reference policy and current behavior policy,
respectively. $\pi_k^t(\cdot\mid o_k^t)$ denotes the candidate
action distribution,
and $\Delta(\mathcal{A}_k)$ is the probability simplex
over agent $k$'s available actions.
The risk-sensitivity coefficient $\beta_s$ balances risk reduction against deviation from the behavior policy. Its normalized exponential form is
\begin{equation}
\pi_k^{\diamondsuit,t}(a\mid o_k^t;\beta_s)=
\frac{\pi_k^\odot(a\mid o_k^t)\exp[-\beta_s\rho_k^t(a)]}
{\sum_{\tilde a\in\mathcal A_k}\pi_k^\odot(\tilde a\mid o_k^t)\exp[-\beta_s\rho_k^t(\tilde a)]}.
\label{eq:safe_reference_explicit}
\end{equation}
This follows by differentiating the objective with a multiplier for probability normalization and normalizing the resulting weights. The safe reference policy is constructed
during training using the context-dependent
risk estimates $\{\rho_k^t(a)\}_{a\in\mathcal A_k}$.

\textbf{Causal Responsibility.}  We quantify agent $k$'s responsibility by the excess tail risk of its factual action relative to the safe reference policy. The positive excess risks are weighted by the corresponding causal relations and averaged over the episode to obtain:
\begin{equation}
\chi_k
=
\frac{
\displaystyle\sum_t\gamma_\chi^t
\sum_{j=1}^{N}\bar{p}_{kj}^t
\left[
\rho_{kj}^t(a_k^t)
-
\mathbb{E}_{
a\sim\pi_k^{\diamondsuit,t}(\cdot\mid o_k^t;\beta_s)
}
\left[\rho_{kj}^t(a)\right]
\right]_+
}{
\displaystyle\sum_t\gamma_\chi^t
},
\label{eq:tail_causal_responsibility}
\end{equation}
where $\chi_k$ denotes agent $k$'s causal responsibility, and $\gamma_\chi\in(0,1]$ is the responsibility discount factor.
$\rho_{kj}^t(a_k^t)$ is the tail risk for agent $j$
under agent $k$'s factual action $a_k^t$.
The operator $[x]_+=\max(x,0)$ retains only positive excess risks. The positive-part operation is applied
before aggregation so that excess risk
to one affected agent is not offset by
risk reductions experienced by other agents. We note that temporal summations are taken over valid decision steps for which the required tail-risk estimates are available.

\section{Responsibility-Guided Preventive Policy Optimization}
\label{subsec:preventive_safety_optimization}
\begin{figure}[pos=!tb]
\centering
\includegraphics[width=0.7\linewidth]{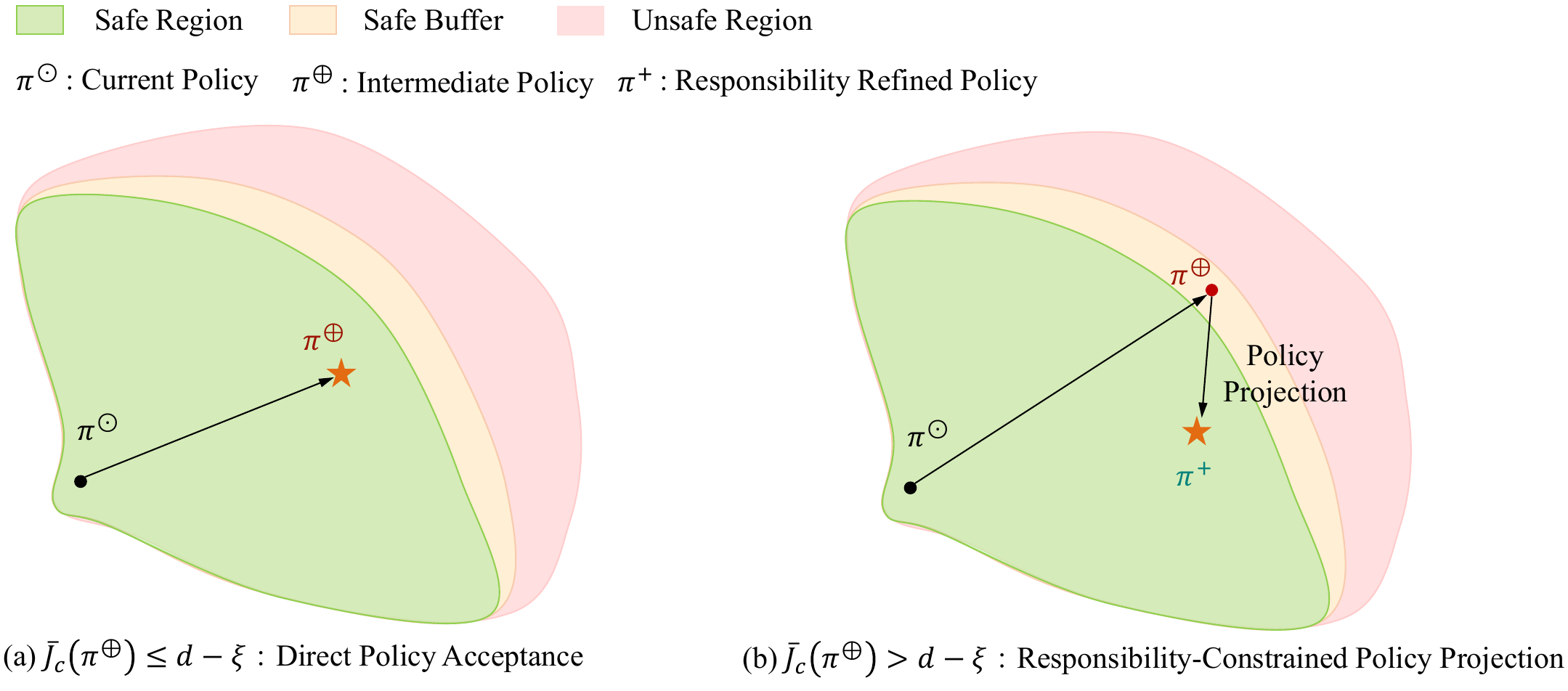}

\caption{Responsibility-guided preventive
policy optimization. An intermediate policy is obtained
by maximizing the shaped reward within a KL trust region.
(a) The policy is accepted if its allowance-adjusted
surrogate safety costs satisfy the reserved margins.
(b) Otherwise, responsibility-based risk-reduction
requirements are imposed during policy projection.}
\label{fig:constrain}

\end{figure}

In this section, we incorporate causal responsibility into preventive policy optimization. Specifically, starting from the current behavior policy $\boldsymbol\pi^\odot$, we optimize an intermediate policy $\boldsymbol\pi^\oplus$ within a KL trust region and evaluate it against surrogate safety constraints with reserved margins. If these constraints are satisfied, the intermediate policy is accepted directly. Otherwise, a constrained projection seeks a corrected policy that remains close to the intermediate policy while satisfying the surrogate safety constraints and responsibility-based risk-reduction requirements, as illustrated in Fig. \ref{fig:constrain}.
\subsection{Responsibility-guided Risk Constraint Formulation}
To translate causal responsibility in \eqref{eq:tail_causal_responsibility} into actionable policy constraints, we determine both the strength and direction of correction for each source agent. We then combine these signals with the reduction capacity permitted by the policy trust region to specify individual risk-reduction requirements, which are imposed as constraints on the candidate policy.

\textbf{Policy Correction Intensity}. To assign stronger corrections to agents with greater causal
responsibility, we map $\chi_k$ to the policy correction
intensity:
\begin{equation}
\alpha_k
=
1+(\alpha_{\max}-1)
\frac{\chi_k}{\chi_k+\sigma_\chi},
\label{eq:responsibility_correction_intensity}
\end{equation}
where $\alpha_k\in[1,\alpha_{\max})$ denotes the policy
correction intensity for agent $k$, $\alpha_{\max}>1$
is its upper bound, and $\sigma_\chi>0$ controls the
responsibility scale. When $\chi_k=0$, $\alpha_k=1$;
as $\chi_k$ increases, $\alpha_k$ approaches
$\alpha_{\max}$, assigning stronger corrections to
agents with greater causal responsibility.

\textbf{Responsibility Gradient}. To guide policy correction toward lower-risk actions, we define the risk advantage function of each action relative to the average risk under the current behavior policy:
\begin{equation}
A_{\chi,k}^{t}(a)
=
\sum_{\tilde a\in\mathcal{A}_k}
\pi_k^\odot(\tilde a\mid o_k^{t})
\rho_k^{t}(\tilde a)-\rho_k^{t}(a),
\label{eq:responsibility_risk_disadvantage}
\end{equation}
where $A_{\chi,k}^{t}(a)$ denotes the risk advantage of action $a$
for agent $k$. A positive value indicates that the action has lower
risk than the behavior-policy average, whereas a negative value
indicates higher risk.
Using the parameterized policy
$\pi_k^t(\cdot\mid o_k^t;\boldsymbol{\theta})$, we compute the
corresponding responsibility gradient:
\begin{equation}
\boldsymbol{g}_{\chi,k}
=
\mathbb{E}_{\mathcal{T}\sim\boldsymbol\pi^{\odot}}
\left[
\sum_t \gamma_\chi^t
\sum_{a\in\mathcal A_k}
\nabla_{\boldsymbol{\theta}}
\pi_k^t(a\mid o_k^t;\boldsymbol{\theta})
A_{\chi,k}^t(a)
\right],
\label{eq:responsibility_gradient}
\end{equation}
where $\boldsymbol{g}_{\chi,k}$ denotes the gradient of the
local risk-advantage surrogate, evaluated at the parameter
values of the current behavior policy while holding the risk
estimates and behavior-policy baseline fixed. A local update
along this gradient increases the surrogate and reduces
the predicted expected risk to first order.

\textbf{Risk-Reduction Requirement.}
The correction intensity $\alpha_k$ determines
the requested reduction $(\alpha_k-1)m_\chi$,
while the responsibility gradient
$\boldsymbol g_{\chi,k}$ identifies a local
direction for reducing risk.
To determine how much reduction the policy update
can support, we maximize the first-order
risk reduction within the local trust region:
\begin{equation}
e_{\chi,k}^{\mathrm{cap}}
=
\max_{\boldsymbol u:\,
\frac12\boldsymbol u^\top\boldsymbol F\boldsymbol u
\leq\delta}
\left\{
\boldsymbol g_{\chi,k}^{\top}\boldsymbol u
\right\},
\label{eq:individual_reduction_capacity}
\end{equation}
where $\boldsymbol u$ denotes a local variation
in the joint-policy parameters,
$\boldsymbol F\succ0$ is the regularized Fisher
information matrix at the current behavior policy,
and $\delta>0$ is the trust-region radius.
The optimal value is
$e_{\chi,k}^{\mathrm{cap}}
=\sqrt{2\delta\,
\boldsymbol g_{\chi,k}^{\top}
\boldsymbol F^{-1}\boldsymbol g_{\chi,k}}$,
by Cauchy--Schwarz after substituting $\boldsymbol w=\boldsymbol F^{1/2}\boldsymbol u$; equality is attained along $\boldsymbol F^{-1}\boldsymbol g_{\chi,k}$ when the gradient is nonzero.

We cap the responsibility-driven request at a
fraction $\kappa_f$ of this capacity, yielding
\begin{equation}
\bar e_{\chi,k}
=
\min
\left\{
(\alpha_k-1)m_\chi,\;
\kappa_f
\sqrt{
2\delta\,
\boldsymbol g_{\chi,k}^{\top}
\boldsymbol F^{-1}\boldsymbol g_{\chi,k}
}
\right\},
\label{eq:capped_responsibility_requirement}
\end{equation}
where $m_\chi>0$ specifies the reduction scale and $\kappa_f\in(0,1]$ is the reduction-capacity scaling coefficient. The resulting requirement accounts for both attributed responsibility and local reduction capacity.

\textbf{Responsibility-guided Risk Constraint.} Imposing uniform risk-reduction requirements without accounting for individual agents' risk responsibility may lead to excessive correction of agents with limited responsibility and insufficient correction of the primary risk sources. To align policy correction with attributed responsibility, we introduce responsibility-guided risk constraints that translate each agent's reduction requirement $\bar e_{\chi,k}$ into an explicit constraint on its candidate policy.  Since the candidate-policy expectation of $A_{\chi,k}^{t}(a)$ measures the predicted risk reduction relative to the behavior policy at the same observation, we require its discounted cumulative value to satisfy:
\begin{equation}
\mathbb{E}_{\mathcal{T}\sim\boldsymbol{\pi}^{\odot}}
\left[
\sum_{t=0}^{T-1}\gamma_{\chi}^{t}
\sum_{a\in\mathcal A_k}
\pi_k(a\mid o_k^t)
A_{\chi,k}^{t}(a)
\right]
\geq\lambda\bar e_{\chi,k},
\qquad k\in\mathcal N_\chi,
\label{eq:source_responsibility_requirement}
\end{equation}
where $\pi_k$ is agent $k$'s component of the
candidate joint policy,
$\mathcal N_\chi=\{k:\bar e_{\chi,k}>0\}$
is the active correction set, and
$\lambda\in(0,1]$ scales the requested reductions. Constraint~\eqref{eq:source_responsibility_requirement}
translates the responsibility-dependent correction
intensity into a minimum improvement that the
candidate policy must achieve.
Its left-hand side evaluates the candidate's
preference for lower-risk actions, while its
right-hand side specifies the required reduction
through $\bar e_{\chi,k}$.
Thus, responsibility determines the correction
requirement, and the risk advantage determines
how changes in action selection satisfy it.

\subsection{Preventive Policy Optimization}
Building on the responsibility-guided risk constraints formulated above, we integrate cooperative policy improvement, safety-cost estimation, and responsibility-guided projection into a preventive update procedure. An intermediate policy is accepted if its estimated safety costs satisfy the reserved margins; otherwise, it is corrected under the safety and responsibility constraints. Starting from the current policy $\boldsymbol{\pi}^{\odot}$, the joint policy is factorized as
$\boldsymbol{\pi}(\boldsymbol a\mid\boldsymbol o)
=\prod_{k=1}^{N}\pi_k(a_k\mid o_k)$, and
the intermediate policy
is obtained by optimizing within
a KL trust region:
\begin{equation}
\begin{aligned}
\boldsymbol{\pi}^{\oplus}
\in\arg\max_{\boldsymbol{\pi}}\quad&
\mathbb{E}_{\boldsymbol{\pi}}
\left[
\sum_{t=0}^{T-1}\gamma_r^t
\sum_{k=1}^{N}r_{k,f}^t
\right]
\\
\mathrm{s.t.}\quad&
\mathbb{E}_{\boldsymbol{\pi}^{\odot}}
\left[
D_{\mathrm{KL}}
\left(
\boldsymbol{\pi}^{\odot}(\cdot\mid\boldsymbol{o})
\,\middle\|\,
\boldsymbol{\pi}(\cdot\mid\boldsymbol{o})
\right)
\right]
\leq\delta,
\end{aligned}
\label{eq:nominal_cooperative_objective}
\end{equation}
where $\boldsymbol{\pi}^{\oplus}$ is the intermediate policy and
$\delta>0$ is the trust-region
radius.

\textbf{Cost Estimation}. To assess the cost of the policy generated during projection, we first formulate a common safety-cost estimator for a generic candidate policy $\boldsymbol{\pi}$. Rather than evaluating this cost directly, we construct a local surrogate relative to the behavior-policy cost. Specifically, we estimate the cost advantage of each joint action under $\boldsymbol{\pi}^{\odot}$ and average these advantages according to $\boldsymbol{\pi}$ at observations collected under the behavior policy. The resulting cost is added to the behavior-policy cost to estimate the candidate policy's safety cost. Specifically, for each agent $k$, the cost advantage under the behavior policy $\boldsymbol\pi^\odot$ is defined as:
\begin{equation}
\begin{aligned}
\widehat A_{C,k}^{\ t,\boldsymbol\pi^\odot}(\boldsymbol a)
={}&
Q_{C,k,\varsigma}^{\boldsymbol\pi^\odot}
(\boldsymbol o^t,\boldsymbol a)
-
\sum_{\widetilde{\boldsymbol a}\in\mathcal A}
\boldsymbol\pi^\odot
(\widetilde{\boldsymbol a}\mid\boldsymbol o^t)
Q_{C,k,\varsigma}^{\boldsymbol\pi^\odot}
(\boldsymbol o^t,\widetilde{\boldsymbol a}),
\end{aligned}
\label{eq:estimated_cost_advantage}
\end{equation}
where
$Q_{C,k,\varsigma}^{\boldsymbol\pi^\odot}$ is the cost
action-value function.  $\widehat A_{C,k}^{\ t,\boldsymbol\pi^\odot}(\boldsymbol a)$ measures how much
the  safety cost of
agent $k$ under joint action $\boldsymbol a$ exceeds
the average over joint actions under the behavior policy. A positive value indicates a higher predicted safety cost than the behavior-policy average, whereas a negative value indicates a lower predicted safety cost. Then, based on the advantage $\widehat A_{C,k}^{\ t,\boldsymbol\pi^\odot}(\boldsymbol a)$, we construct
a local surrogate for the safety cost of the candidate policy $\boldsymbol\pi$:
\begin{equation}
\begin{aligned}
\widehat J_{C,k}(\boldsymbol\pi)
={}&
J_{C,k}(\boldsymbol\pi^\odot)
+
\mathbb E_{\mathcal{T}\sim\boldsymbol\pi^\odot}
\left[
\sum_{t=0}^{T-1}\gamma_c^t
\sum_{\boldsymbol a\in\mathcal A}
\boldsymbol\pi(\boldsymbol a\mid\boldsymbol o^t)
\widehat A_{C,k}^{t,\boldsymbol\pi^\odot}(\boldsymbol a)
\right],
\end{aligned}
\label{eq:critic_cost_estimation}
\end{equation}
where
$J_{C,k}(\boldsymbol\pi^\odot)=\mathbb E_{\boldsymbol\pi^\odot}
[\sum_{t=0}^{T-1}\gamma_c^t c_k^t]$
is the discounted safety cost under the behavior
policy. $\widehat J_{C,k}(\boldsymbol\pi)$ is the evaluated safety cost of the candidate policy. $\mathcal{T}$ denotes a trajectory generated under the current behavior policy $\boldsymbol\pi^\odot$.

\textbf{Surrogate Safety Constraint.} We obtain an estimate of the safety cost for the candidate policy via \eqref{eq:critic_cost_estimation}; however, approximation errors in the cost action-value function and shifts in the policy distribution may lead to an underestimation of this cost. To account for this uncertainty, we introduce a non-negative margin $\epsilon_k$ for each agent and impose the surrogate safety constraint:
\begin{equation}
\bar J_{C,k}(\boldsymbol\pi)
=
\hat J_{C,k}(\boldsymbol\pi)+\epsilon_k
\leq d_k-\xi_k,
\qquad k=1,\ldots,N,
\label{eq:allowance_adjusted_safety_constraint}
\end{equation}
where $\hat J_{C,k}(\boldsymbol\pi)$ is the estimated safety cost under the candidate policy $\boldsymbol\pi$ and $\bar J_{C,k}(\boldsymbol\pi)$ is the corresponding allowance-adjusted cost. $\epsilon_k$ is selected empirically to compensate for cost underestimation. $d_k$ is the safety-cost limit, and $\xi_k>0$ is the reserved safety margin. The intermediate policy is accepted when $\bar J_{C,k}(\boldsymbol\pi^\oplus)\le d_k-\xi_k$ for every agent; otherwise, preventive projection is activated. Algorithm~\ref{alg:preventive_policy_optimization} summarizes candidate verification and fallback.

Let $\mathcal B_\delta(\boldsymbol\pi^\odot)$ denote the set of policies satisfying the KL constraint in \eqref{eq:nominal_cooperative_objective}.\par
\textbf{Preventive Safety Projection Problem}. Combining the surrogate safety constraint and responsibility-guided risk constraint, the final preventive safety projection problem can be formulated as:
\begin{equation}
\label{eq:responsibility_constrained_optimization}
\begin{aligned}
\boldsymbol{\pi}^{+}
\in\operatorname*{arg\,min}_{
\boldsymbol{\pi}\in
\mathcal{B}_{\delta}(\boldsymbol{\pi}^{\odot})}
\quad&
\mathbb{E}_{\mathcal{T}\sim\boldsymbol\pi^\odot}
\left[
D_{\mathrm{KL}}
\left(
\boldsymbol{\pi}^{\oplus}(\cdot\mid\boldsymbol{o})
\,\middle\|\,
\boldsymbol{\pi}(\cdot\mid\boldsymbol{o})
\right)
\right]
\\
\mathrm{s.t.}\quad&
\bar{\boldsymbol{J}}_C(\boldsymbol{\pi})
\preceq\boldsymbol{d}-\boldsymbol{\xi},
\\
&
\mathbb{E}_{\mathcal{T}\sim\boldsymbol\pi^\odot}
\left[
\sum_{t=0}^{T-1}\gamma_{\chi}^{t}
\sum_{a\in\mathcal A_k}
\pi_k(a\mid o_k^t)
A_{\chi,k}^{t}(a)
\right]
\geq\lambda\bar e_{\chi,k},
\qquad k\in\mathcal N_\chi.
\end{aligned}
\end{equation}
where
$\bar{\boldsymbol J}_C(\boldsymbol\pi)
=[\bar J_{C,1}(\boldsymbol\pi),\ldots,
\bar J_{C,N}(\boldsymbol\pi)]^\top$,
$\boldsymbol d=[d_1,\ldots,d_N]^\top$, and
$\boldsymbol\xi=[\xi_1,\ldots,\xi_N]^\top$
collect the allowance-adjusted safety costs,
cost limits, and reserved safety margins,
respectively.
The objective in \eqref{eq:responsibility_constrained_optimization}
preserves the intermediate cooperative policy
by minimizing policy divergence.
Constraint~\eqref{eq:allowance_adjusted_safety_constraint}
enforces the reserved margins in the safety costs, while
\eqref{eq:source_responsibility_requirement}
requires the responsibility-guided risk reductions. For each $\lambda \in \{\lambda_1,\ldots,\lambda_n\}$, a local approximation of
\eqref{eq:responsibility_constrained_optimization}
is solved, and the candidate is verified against
the surrogate safety-cost, responsibility, and
KL constraints in the original formulation.
The first scale yielding a verified candidate is
denoted by $\lambda^*$.
If no scale yields a verified candidate, the
responsibility constraints are removed and only
the safety-cost constraints are retained within
the same trust region.
If this fallback also fails verification, the
update is rejected and
$\boldsymbol{\pi}^{\odot}$ is retained. Algorithm~\ref{alg:preventive_policy_optimization} summarizes the process.

\begin{algorithm}[t]
\caption{Responsibility-Guided Preventive Policy Optimization}
\label{alg:preventive_policy_optimization}
\small
\textbf{Input:} Behavior policy $\boldsymbol\pi^\odot$, trajectory batch $\mathcal D$, cost critics $\{Q_{C,k,\varsigma}^{\boldsymbol\pi^\odot}\}_{k=1}^N$, cost limits $\boldsymbol d$, reserved margins $\boldsymbol\xi$, KL radius $\delta$, and candidate scales $\Lambda$.\par
\textbf{Output:} A verified candidate, or $\boldsymbol\pi^\odot$ if the update is rejected.
\begin{enumerate}[label=\arabic*:,leftmargin=2em,itemsep=1pt,topsep=4pt]
\item Obtain $\boldsymbol\pi^\oplus$ using \eqref{eq:nominal_cooperative_objective}.
\item Evaluate $\bar{\boldsymbol J}_C(\boldsymbol\pi^\oplus)$ using \eqref{eq:critic_cost_estimation} and \eqref{eq:allowance_adjusted_safety_constraint}.
\item \textbf{If} $\bar{\boldsymbol J}_C(\boldsymbol\pi^\oplus)\preceq\boldsymbol d-\boldsymbol\xi$ and $\boldsymbol\pi^\oplus\in\mathcal B_\delta(\boldsymbol\pi^\odot)$, \textbf{return} $\boldsymbol\pi^\oplus$.
\item Compute $\bar e_{\chi,k}$ using \eqref{eq:capped_responsibility_requirement}; set $\mathcal N_\chi=\{k:\bar e_{\chi,k}>0\}$.
\item \textbf{If} $\mathcal N_\chi\ne\varnothing$, examine $\lambda\in\Lambda$ in descending order:
\begin{enumerate}[label=\alph*.,leftmargin=1.7em,itemsep=1pt,topsep=1pt]
\item Solve the local approximation of \eqref{eq:responsibility_constrained_optimization} at scale $\lambda$ to obtain a candidate $\boldsymbol\pi$.
\item If the candidate satisfies all original surrogate safety-cost, responsibility, and KL constraints, set $\lambda^*=\lambda$ and \textbf{return} $\boldsymbol\pi$.
\end{enumerate}
\item Attempt projection without responsibility constraints.
\item If its candidate satisfies the original surrogate safety-cost and KL constraints, \textbf{return} that candidate; otherwise \textbf{return} $\boldsymbol\pi^\odot$.
\end{enumerate}
\end{algorithm}

\section{Theoretical Analysis}
\label{sec:theoretical_analysis}
This section establishes monotonicity of the safe reference policy's predicted risk and sufficient local conditions for non-degrading environmental return.

\begin{assumption}[Policy and Risk Regularity]
\label{ass:policy_risk_regular}
For each agent $i$ and decision step $t$, the action set $\mathcal{A}_i$ is finite, the behavior policy has full support,
$\pi_i^{\odot}(a\mid o_i^t)>0$, and the aggregated tail risk $\rho_i^t(a)$ is finite for every $a\in\mathcal{A}_i$.
\end{assumption}

\begin{definition}[Expected Risk]
\label{def:risk_partition}
For each agent $i$, decision step $t$, and risk-sensitivity coefficient
$\beta_s\ge0$, define the expected aggregated tail risk under the safety
reference policy $\pi_i^{\diamondsuit,t}(a\mid o_i^t;\beta_s)$ in
\eqref{eq:unified_safe_reference} as
$
R_i^t(\beta_s)
=
\mathbb{E}_{a\sim\pi_i^{\diamondsuit,t}(\cdot\mid o_i^t;\beta_s)}
\left[
\rho_i^t(a)
\right].
$
In particular, when $\beta_s=0$, the safe reference policy reduces to
the current behavior policy $\pi_i^{\odot}$, and thus
$
R_i^t(0)
=
\mathbb{E}_{a\sim\pi_i^\odot (\cdot\mid o_i^t)}
\left[
\rho_i^t(a)
\right].
$ In addition, define the partition function as
$
Z_i^t(\beta_s)
=
\sum_{a\in\mathcal{A}_i}
\pi_i^{\odot}(a\mid o_i^t)
\exp[-\beta_s\rho_i^t(a)]
$.
\end{definition}

\begin{proposition}[Risk Monotonicity]
\label{prop:risk_monotonicity}
Under Assumption~\ref{ass:policy_risk_regular}, the expected aggregated
tail risk $R_i^t(\beta_s)$ is non-increasing with respect to the
risk-sensitivity coefficient $\beta_s$. In particular,
\begin{equation}
\frac{\partial R_i^t(\beta_s)}
{\partial \beta_s}
=
-
\operatorname{Var}_{a\sim\pi_i^{\diamondsuit,t}(\cdot;\beta_s)}
\left[
\rho_i^t(a)
\right]
\le 0.
\label{eq:risk_monotonicity}
\end{equation}
Moreover, the inequality is strict whenever
$\rho_i^t(a)$ is nonconstant over the support of
$\pi_i^{\diamondsuit,t}(\cdot;\beta_s)$.
\end{proposition}

\begin{proof}
From Definition~\ref{def:risk_partition}, the log-partition function
satisfies
\begin{equation}
\frac{\partial}{\partial\beta_s}
\log Z_i^t(\beta_s)
=
-R_i^t(\beta_s),
\qquad
\frac{\partial^2}{\partial\beta_s^2}
\log Z_i^t(\beta_s)
=
\operatorname{Var}_{a\sim
\pi_i^{\diamondsuit,t}(\cdot\mid o_i^t;\beta_s)}
[\rho_i^t(a)].
\label{eq:log_partition_risk_relation}
\end{equation}
Therefore,
$\partial R_i^t(\beta_s)/\partial\beta_s$
equals the negative risk variance and is nonpositive, with strict
inequality whenever the variance is positive.

\end{proof}

These statements concern predicted aggregated tail risk with the fitted risk estimates held fixed. They do not imply monotonic improvement of realized safety costs as $\beta_s$ increases.

\begin{assumption}[Local Projection and Certification Regularity]
\label{ass:responsibility_projection_regular}
For a fixed policy update, let $\boldsymbol{x}$ denote the local policy parameters and let $\boldsymbol{F}\succ\boldsymbol{0}$ be a fixed regularized Fisher information matrix, with $\|\boldsymbol{v}\|_{\boldsymbol{F}}= \sqrt{\boldsymbol{v}^{\top}\boldsymbol{F}\boldsymbol{v}}$. The trust-region radius satisfies $\delta>0$, and $\bar e_{\chi,i}\ge0$ for every agent $i$. On a nonempty interval $\mathcal I\subseteq[0,1]$, the local quadratic projection is feasible, its active affine constraints remain unchanged and have linearly independent gradients, and the Fisher trust-region boundary is inactive. The environmental team return $J_R$ is $L_R$-Lipschitz
under the Fisher norm on a neighborhood containing
the intermediate policy and the projected policies
considered below, with $L_R>0$.
The intermediate policy satisfies
$
J_R(\boldsymbol{\pi}^{\oplus})
-
J_R(\boldsymbol{\pi}^{\odot})
\ge
G^{\oplus}-\varepsilon^R,
$
where $G^{\oplus}$ is the estimated environmental-return improvement and $\varepsilon^R\ge0$ is its one-sided approximation-error allowance. For safety certification, let $\mathcal E_C$ denote
the event on which
$
J_{C,i}(\boldsymbol\pi)
\le
\bar J_{C,i}(\boldsymbol\pi)
=
\widehat J_{C,i}(\boldsymbol\pi)+\epsilon_i
$
holds simultaneously for every agent and every
candidate policy that may be selected during the update.
\end{assumption}

\begin{proposition}[Responsibility-Induced Improvement Bound]
\label{prop:intervention_improvement_coupling}
Under Assumption~\ref{ass:responsibility_projection_regular},
consider $\lambda\in\mathcal I$ and the
local Fisher-quadratic projection with fixed affine
active constraints. Writing its solution as
$\boldsymbol x^+(\lambda)=\boldsymbol x_0+\lambda\boldsymbol v$
and denoting the intermediate policy parameters by
$\boldsymbol x^\oplus$, define
$\varphi_2=\|\boldsymbol v\|_{\boldsymbol F}^2$,
$\varphi_1=(\boldsymbol x_0-\boldsymbol x^{\oplus})^{\top}
\boldsymbol F\boldsymbol v$, and
$\varphi_0=\|\boldsymbol x_0-\boldsymbol x^\oplus\|_{\boldsymbol F}^2$.
Then the local squared displacement is
$\varphi_2\lambda^2+2\varphi_1\lambda+\varphi_0$.
The original environmental-return improvement obeys
\begin{equation}
\begin{aligned}
J_R(\boldsymbol{\pi}^{+})
-
J_R(\boldsymbol{\pi}^{\odot})
\ge{}&G^{\oplus}-\varepsilon^{R}-L_R\sqrt{\varphi_2\lambda^2
+2\varphi_1\lambda+\varphi_0}.
\end{aligned}
\label{eq:intervention_improvement_bound}
\end{equation}
Here $G^\oplus-\varepsilon^R$ is an assumed valid lower
bound on the nominal \emph{environmental} return
improvement. The coefficients $\varphi_0,\varphi_1,
\varphi_2$ describe only the fixed-active-set local
Fisher approximation; the inequality is not an
unconditional global performance guarantee.
\end{proposition}

\begin{proof}
Write the fixed active constraints as $\boldsymbol C\boldsymbol x=\boldsymbol r-\lambda\boldsymbol e$, with full-row-rank $\boldsymbol C$. Since the trust-region boundary is inactive, stationarity of the Fisher-quadratic projection gives
\begin{equation}
\boldsymbol x^+(\lambda)=\boldsymbol x^\oplus-\boldsymbol F^{-1}\boldsymbol C^\top
(\boldsymbol C\boldsymbol F^{-1}\boldsymbol C^\top)^{-1}
(\boldsymbol C\boldsymbol x^\oplus-\boldsymbol r+\lambda\boldsymbol e).
\end{equation}
The solution is affine in $\lambda$. Expanding its squared Fisher distance from $\boldsymbol x^\oplus$ gives $\varphi_2\lambda^2+2\varphi_1\lambda+\varphi_0$. The Lipschitz assumption bounds the return change from $\boldsymbol\pi^\oplus$ to $\boldsymbol\pi^+$ below by minus $L_R$ times this distance. Adding the assumed bound $G^\oplus-\varepsilon^R$ proves the result.
\end{proof}

\begin{theorem}[Sufficient Local Range for Responsibility Enforcement]
\label{thm:certified_responsibility_range}
Under Assumption~\ref{ass:responsibility_projection_regular},
consider a local enforcement interval satisfying the
intervention geometry established in
Proposition~\ref{prop:intervention_improvement_coupling}.
If $G^{\oplus}>\varepsilon^R$,
$\varphi_0\le[(G^{\oplus}-\varepsilon^R)/L_R]^2$,
and $\varphi_2>0$, then the responsibility enforcement parameter has the following sufficient
non-degradation range within the local interval
$\mathcal I$. Define
$b_R=(G^\oplus-\varepsilon^R)/L_R>0$; then
\begin{equation}
\begin{aligned}
\lambda&\in\mathcal I\cap[0,\lambda^{\dagger}],\quad \lambda^{\dagger}
=\min\Biggl\{1,\,
\frac{\sqrt{\varphi_1^2+
\varphi_2(b_R^2-\varphi_0)}-\varphi_1}
{\varphi_2}\Biggr\}.
\end{aligned}
\label{eq:certified_responsibility_range}
\end{equation}
For each $\lambda\in\mathcal I\cap[0,\lambda^\dagger]$,
let $\boldsymbol\pi^+(\lambda)$ be the policy corresponding
to the local quadratic projection solution characterized
in Proposition~\ref{prop:intervention_improvement_coupling}.
Its environmental return is non-degrading:
$
J_R(\boldsymbol\pi^+(\lambda))
\ge J_R(\boldsymbol\pi^\odot).
$
If this policy also passes verification of the original
surrogate safety-cost constraints,
$
\bar J_{C,i}(\boldsymbol\pi^+(\lambda))
\le d_i-\xi_i
$
for every agent $i$, then, on $\mathcal E_C$,
it satisfies
$
J_{C,i}(\boldsymbol\pi^+(\lambda))
\le d_i-\xi_i.
$
If $\mathbb P(\mathcal E_C)\ge1-\alpha_C$,
this safety implication holds with probability at least
$1-\alpha_C$ for the considered update.
\end{theorem}

\begin{proof}
By Proposition~\ref{prop:intervention_improvement_coupling}, non-degradation holds if
$\varphi_2\lambda^2+2\varphi_1\lambda+\varphi_0\le b_R^2$.
The stated conditions make its nonnegative root
$[\sqrt{\varphi_1^2+\varphi_2(b_R^2-\varphi_0)}-\varphi_1]/\varphi_2$.
Intersecting the resulting range with $[0,1]$ and $\mathcal I$ gives the claim. For a verified candidate, on $\mathcal E_C$,
$J_{C,i}(\boldsymbol\pi^+)\le\bar J_{C,i}(\boldsymbol\pi^+)\le d_i-\xi_i$.
The probability statement follows from the assumed bound on $\mathbb P(\mathcal E_C)$. No safety guarantee is asserted for an unverified fallback policy.
\end{proof}

\begin{remark}
Responsibility enforcement balances source-specific risk reduction with preservation of environmental return. The derived bound quantifies this trade-off through the projection geometry and reward sensitivity, while $\lambda^{\dagger}$ provides a sufficient upper limit on enforcement strength within the local validity interval.
\end{remark}

The range above is a sufficient theoretical condition, not an upper bound computed by Algorithm~\ref{alg:preventive_policy_optimization}. The empirical allowance $\epsilon_i$ does not itself establish the event $\mathcal E_C$ or its probability.

\section{Experiment Setup}
\label{sec:exp_setup}
\subsection{Simulation Settings}
The simulation environment is implemented using a modified version of the \textit{highway-env} simulator~\cite{Edouard2018Highway}. The main road is 520~m long, with a 100~m merging section beginning 320~m downstream from the origin. Further details of the simulation scenario can be found in \cite{teng2026tits}. Vehicle motion is simulated using a kinematic bicycle model~\cite{Polack2017IV}, and all vehicles are 5.0~m long and 2.0~m wide. For HDVs, longitudinal motion and lane-changing behavior are modeled by the Intelligent Driver Model~\cite{treiber2000congested} and the Minimizing Overall Braking Induced by Lane Changes model~\cite{2001Preferred}, respectively. To emulate variability in human driving behavior, 5\% random perturbations are added to their acceleration and steering commands. For CAVs, high-level decisions are generated by the learned policy and executed by low-level PID controllers at 5~Hz.
At the beginning of each episode, CAVs and HDVs are randomly assigned to predefined spawn points within the upstream 0--220~m region of the main road and ramp, with equal numbers of vehicles initialized on each. Initial positions are perturbed by uniformly distributed noise within $\pm 1.5$~m, and initial speeds are sampled uniformly from 20 to 30~m/s. To evaluate the robustness of the proposed method under varying traffic densities, experiments are conducted at three density levels: \textit{Low}, with 2--3 CAVs and 2--3 HDVs; \textit{Medium}, with 3--4 CAVs and 3--4 HDVs; and \textit{High}, with 4--5 CAVs and 4--5 HDVs.

\subsection{Implementation Details}
To thoroughly evaluate the proposed framework, we compare CARGO against representative baselines spanning four categories: causal-aware MARL, ramp-merging-specific algorithms, general MARL, and traditional methods. Specifically, the causal-aware methods include Situation-dependent Causal Influence-based Cooperative multi-agent reinforcement learning (SCIC)~\cite{du2024situation} and Causality Aware~\cite{cao2026causality}. The ramp-merging-specific algorithms include Safe Multi-Agent Advantage Actor-Critic (Safe-MAA2C)~\cite{Chendong2023TITS} and Multi-Agent Proximal Policy Optimization with Prior Intent Sharing (MAPPO-PIS)~\cite{guo2024ECCV}. The general MARL baselines comprise Multi-Agent Proximal Policy Optimization (MAPPO)~\cite{yu2022mappo} and Multi-Agent Actor-Critic using Kronecker-factored Trust Region (MAACKTR)~\cite{wu2017MAACKTR}. Additionally, Model Predictive Control (MPC)~\cite{Cao2013MPC} serves as a traditional baseline.
All learning-based methods use the same base observation and action spaces, environment reward, and safety cost definitions, while preserving their algorithm-specific mechanisms. All methods, including MPC, are evaluated on the same test scenarios with identical traffic demand, vehicle initialization, and episode termination rules.

The evaluation metrics include: 1) \textit{Success Rate}, defined as the proportion of collision-free episodes in which all initial ramp vehicles complete merging; 2) \textit{Crash Rate}, the proportion of episodes with any collision; 3) \textit{Merging Time}, the time from the first ramp vehicle entering the acceleration lane to the last completing its merge, measured only in successful episodes; 4) \textit{Average Velocity}, the mean speed across all vehicles and time steps; 5) \textit{Min TTC}, the episode minimum TTC across longitudinal and lateral conflicts, averaged over episodes; and 6) \textit{TTC Violation Rate}, the proportion of decision steps with TTC below the corresponding safety threshold.

All learning-based methods are trained for $2 \times 10^{6}$ environment steps with a common initial learning rate of $5 \times 10^{-4}$. Training is repeated across three independent random seeds. In all training plots, curves show the mean across the three runs, and shaded regions indicate standard deviation. Reported rewards are environmental rewards and exclude auxiliary intrinsic rewards. For testing, each traffic-density setting is evaluated using three independent random seeds, with 100 distinct test episodes per seed. All experiments are conducted on a workstation equipped with an Intel Core i9-13900HX CPU. For latent-context inference and causal graph construction, the history window is set to $H_w=10$, the latent dimension to $l_z=32$, and the posterior sample size to $M=64$, with the normalized strength and confidence thresholds fixed at $0.5$ and $0.8$, respectively. For tail-risk estimation, the diffusion model predicts 5 future steps using 100 diffusion steps and 10 DDIM sampling steps. CVaR is estimated from $L=8$ response samples at a confidence level of $\beta=0.75$. For causal contribution and responsibility assessment, we set the causal contribution reward weight to $\eta_c=1$, the risk sensitivity coefficient to $\beta_s=1$, the responsibility scaling parameter to $\sigma_\chi=1$, and the discount factor to $\gamma_\chi=0.99$. For responsibility-guided correction, we set the maximum intensity to $\alpha_{\max}=2.0$, the risk-reduction scale to $m_\chi=2$, and the capacity scaling coefficient to $\kappa_f=0.5$, with candidate scales $\Lambda={1.0,0.75,0.5,0.25,0.125}$. For preventive safety optimization, we use a safety-cost limit of $d_k=75$, an approximation-error allowance of $\epsilon_k=5$, and a reserved safety margin of $\xi_k=5$ for all agents, with a KL trust-region radius of $\delta=0.1$.

\section{Experimental Results and Analysis}
\label{sec:results}

\begin{figure}[pos=!tb]
\centering
\includegraphics[width=1\linewidth]{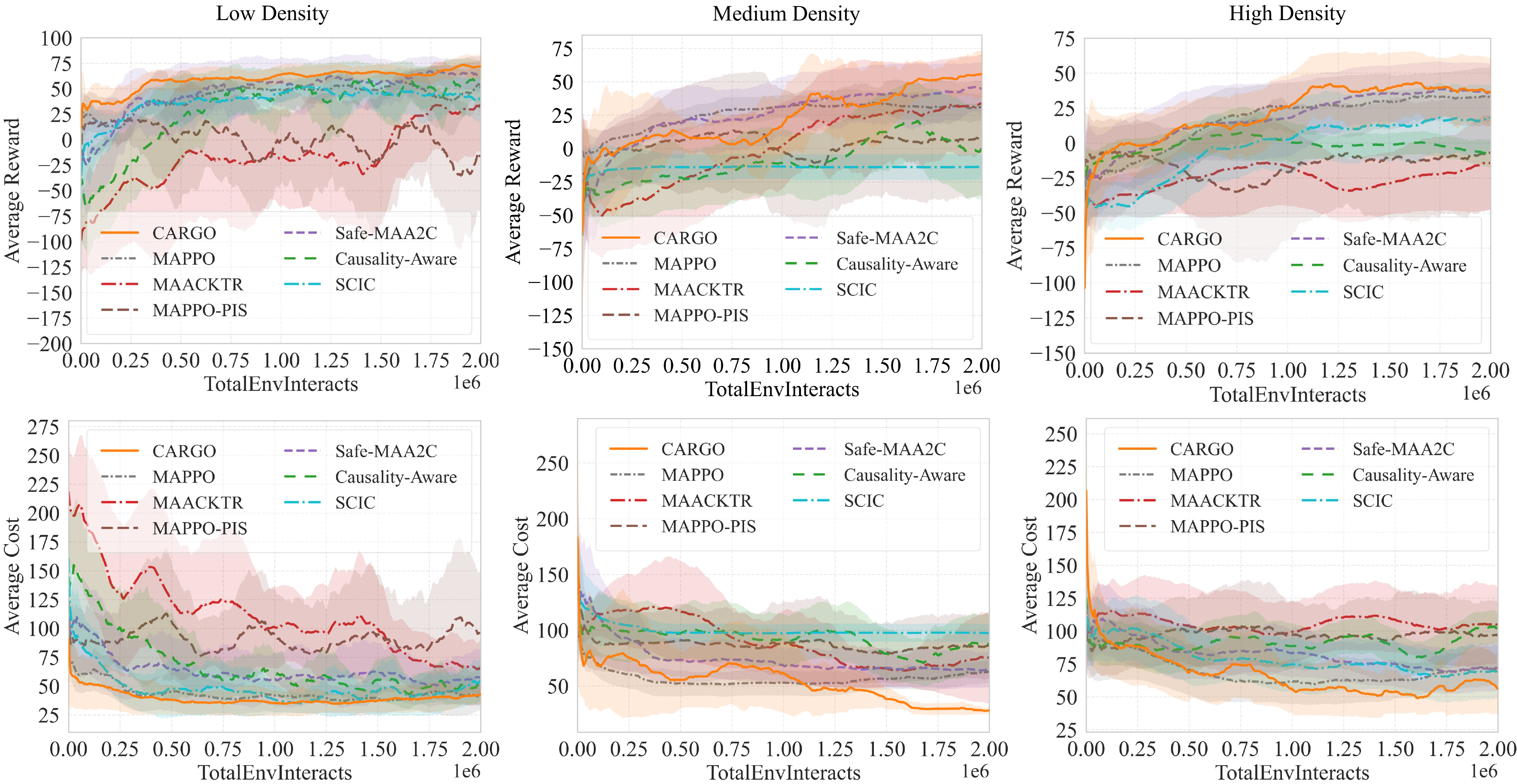}

\caption{Comparative study under low-, medium-, and high-density traffic. CARGO achieves higher or comparable rewards while generally maintaining lower costs than the baselines, demonstrating a favorable balance between task performance and safety.}
\label{fig:comparative_single}
\end{figure}

\begin{table*}[!t]
\centering
\caption{Comparative study under different traffic densities.}
\label{tab:comparative_density}
\scriptsize
\renewcommand{\arraystretch}{1.3}
\setlength{\tabcolsep}{2pt}
\setlength{\heavyrulewidth}{0.55pt}
\setlength{\lightrulewidth}{0.35pt}
\setlength{\cmidrulewidth}{0.30pt}
\begin{tabular*}{\textwidth}{@{\extracolsep{\fill}}llcccccc@{}}
\toprule
\multirow{2}{*}{\shortstack[c]{Traffic\\Density}}
& \multicolumn{1}{c}{\multirow{2}{*}{Method}}
& Success Rate
& Crash Rate
& Merging Time
& Average Velocity
& Min TTC
& TTC Violation Rate \\
&
&
(\%) $\uparrow$
& (\%) $\downarrow$
& (s) $\downarrow$
& (m/s) $\uparrow$
& (s) $\uparrow$
& (\%) $\downarrow$ \\
\midrule

\multirow{6}{*}{\shortstack[c]{Low-density\\Traffic}}
& SCIC~\cite{du2024situation}
& $60 \pm 1.73$ & $22.67 \pm 4.04$ & $10.83 \pm 0.66$
& $21.96 \pm 0.22$ & $1.86 \pm 0.11$ & $3.99 \pm 0.43$ \\
& Causality Aware~\cite{cao2026causality}
& $79.33 \pm 3.21$ & $15.67 \pm 3.51$
& $\underline{9.53 \pm 0.06}$
& $\underline{24.80 \pm 0.25}$
& $\underline{3.33 \pm 0.24}$
& $\underline{0.60 \pm 0.23}$ \\
& MAACKTR~\cite{wu2017MAACKTR}
& $58.67 \pm 4.04$ & $30.67 \pm 3.21$ & $11.26 \pm 0.25$
& $23.47 \pm 0.11$ & $1.89 \pm 0.04$ & $5.90 \pm 0.58$ \\
& MAPPO~\cite{yu2022mappo}
& $78.67 \pm 3.06$ & $10.67 \pm 3.06$ & $10.57 \pm 0.23$
& $23.22 \pm 0.19$ & $2.99 \pm 0.11$ & $1.67 \pm 2.05$ \\
& Safe-MAA2C~\cite{Chendong2023TITS}
& $\underline{83.33 \pm 3.51}$
& $\underline{2.33 \pm 1.53}$
& $10.08 \pm 0.12$
& $23.99 \pm 0.10$ & $2.90 \pm 0.12$ & $1.44 \pm 0.23$ \\
& MAPPO-PIS~\cite{guo2024ECCV}
& $58.67 \pm 5.51$ & $30.67 \pm 9.07$ & $10.55 \pm 0.41$
& $23.98 \pm 0.18$ & $2.08 \pm 0.23$ & $5.74 \pm 1.47$ \\
& MPC~\cite{Cao2013MPC}
& $73.67 \pm 8.62$ & $16.67 \pm 5.86$ & $10.11 \pm 0.10$
& $24.75 \pm 0.16$ & $2.32 \pm 0.07$ & $4.98 \pm 1.24$ \\
& \textbf{CARGO (ours)}
& $\boldsymbol{88.00 \pm 1.00}$
& $\boldsymbol{2.00 \pm 1.00}$
& $\boldsymbol{9.47 \pm 0.21}$
& $\boldsymbol{24.85 \pm 0.08}$
& $\boldsymbol{3.70 \pm 0.22}$
& $\boldsymbol{0.48 \pm 0.05}$ \\

\midrule

\multirow{6}{*}{\shortstack[c]{Medium-density\\Traffic}}
& SCIC~\cite{du2024situation}
& $20.00 \pm 1.73$ & $15.67 \pm 2.52$ & $12.11 \pm 0.22$
& $18.83 \pm 0.11$ & $0.50 \pm 0.12$ & $4.32 \pm 0.16$ \\
& Causality Aware~\cite{cao2026causality}
& $28.00 \pm 1.00$
& $\underline{5.67 \pm 1.53}$
& $12.86 \pm 0.69$
& $18.11 \pm 0.02$
& $\underline{2.24 \pm 0.17}$
& $\underline{2.73 \pm 0.61}$ \\
& MAACKTR~\cite{wu2017MAACKTR}
& $13.67 \pm 1.53$ & $56.67 \pm 3.21$ & $12.05 \pm 0.33$
& $21.47 \pm 0.14$ & $0.75 \pm 0.12$ & $12.15 \pm 1.48$ \\
& MAPPO~\cite{yu2022mappo}
& $\underline{48.33 \pm 1.15}$
& $13.33 \pm 4.51$ & $11.87 \pm 0.01$
& $21.42 \pm 0.05$ & $2.13 \pm 0.08$ & $3.10 \pm 0.34$ \\
& Safe-MAA2C~\cite{Chendong2023TITS}
& $36.00 \pm 3.46$ & $8.00 \pm 2.65$
& $\boldsymbol{11.19 \pm 0.26}$
& $22.54 \pm 0.07$ & $1.66 \pm 0.17$ & $3.46 \pm 0.36$ \\
& MAPPO-PIS~\cite{guo2024ECCV}
& $23.67 \pm 5.03$ & $56.00 \pm 7.00$ & $11.31 \pm 0.41$
& $22.64 \pm 0.10$ & $1.02 \pm 0.16$ & $10.69 \pm 0.57$ \\
& MPC~\cite{Cao2013MPC}
& $23.00 \pm 5.29$ & $34.00 \pm 10.00$ & $11.41 \pm 0.34$
& $\boldsymbol{23.79 \pm 0.18}$
& $1.01 \pm 0.03$ & $8.98 \pm 0.82$ \\
& \textbf{CARGO (ours)}
& $\boldsymbol{50.67 \pm 1.53}$
& $\boldsymbol{2.67 \pm 0.58}$
& $\underline{11.24 \pm 0.32}$
& $\underline{22.83 \pm 0.10}$
& $\boldsymbol{2.47 \pm 0.16}$
& $\boldsymbol{1.38 \pm 0.43}$ \\

\midrule

\multirow{8}{*}{\shortstack[c]{High-density\\Traffic}}
& SCIC~\cite{du2024situation}
& $12.33 \pm 0.58$ & $39.33 \pm 4.16$ & $13.53 \pm 1.24$
& $19.03 \pm 0.08$ & $1.14 \pm 0.03$ & $6.61 \pm 0.27$ \\
& Causality Aware~\cite{cao2026causality}
& $12.67 \pm 2.08$ & $33.33 \pm 2.31$ & $13.08 \pm 1.89$
& $17.12 \pm 0.12$ & $1.16 \pm 0.11$ & $5.32 \pm 0.21$ \\
& MAACKTR~\cite{wu2017MAACKTR}
& $11.33 \pm 1.15$ & $73.00 \pm 4.58$ & $12.90 \pm 3.63$
& $20.34 \pm 0.06$ & $0.31 \pm 0.06$ & $15.82 \pm 0.93$ \\
& MAPPO~\cite{yu2022mappo}
& $12.33 \pm 2.89$ & $11.00 \pm 2.00$ & $12.95 \pm 0.97$
& $18.69 \pm 0.05$
& $\underline{1.59 \pm 0.11}$
& $\underline{2.92 \pm 0.18}$ \\
& Safe-MAA2C~\cite{Chendong2023TITS}
& $18.00 \pm 3.46$
& $\underline{10.67 \pm 2.31}$
& $13.64 \pm 0.37$
& $21.09 \pm 0.07$ & $1.15 \pm 0.02$ & $5.09 \pm 0.21$ \\
& MAPPO-PIS~\cite{guo2024ECCV}
& $15.33 \pm 3.21$ & $41.67 \pm 5.51$
& $\underline{12.76 \pm 0.63}$
& $20.16 \pm 0.03$ & $1.10 \pm 0.16$ & $7.63 \pm 0.96$ \\
& MPC~\cite{Cao2013MPC}
& $\underline{18.67 \pm 2.08}$
& $45.00 \pm 1.73$ & $14.03 \pm 1.08$
& $\underline{21.80 \pm 0.01}$
& $0.63 \pm 0.05$ & $10.65 \pm 0.79$ \\
& \textbf{CARGO (ours)}
& $\boldsymbol{31.00 \pm 3.61}$
& $\boldsymbol{4.67 \pm 2.89}$
& $\boldsymbol{12.74 \pm 0.81}$
& $\boldsymbol{21.97 \pm 0.14}$
& $\boldsymbol{1.89 \pm 0.03}$
& $\boldsymbol{2.37 \pm 0.11}$ \\

\bottomrule
\end{tabular*}
\begin{minipage}{\textwidth}
\scriptsize
\rmfamily
\justifying
\vspace{1mm}
\noindent
Note: The best and second-best results for each metric are highlighted in \textbf{bold} and \underline{underlined}, respectively.
\par
\end{minipage}

\end{table*}

\subsection{Comparative Study}
Table~\ref{tab:comparative_density} compares performance across three traffic densities. CARGO achieves the highest success rate, lowest crash rate, and largest minimum TTC in each setting. Relative to the strongest baseline for each metric, crash rates decrease by 14.16\%, 52.91\%, and 56.23\% under low-, medium-, and high-density traffic. High-density success increases from 18.67\% to 31.00\%, a relative improvement of 66.04\%. Many high-density episodes terminate without a collision but without successful merging, so crash rate and success rate describe distinct outcomes. CARGO also achieves the highest average velocity and shortest merging time at low and high density; medium-density efficiency remains competitive.

Fig.~\ref{fig:comparative_single} shows the corresponding learning curves. CARGO maintains competitive environmental rewards with comparatively low late-training costs, particularly as traffic density increases. The evaluation therefore considers task completion and efficiency together with collision avoidance.

\subsection{Ablation Study}

\begin{table*}[!t]
\centering
\caption{Ablation study under different traffic densities.}
\label{tab:ablation_density}
\scriptsize
\renewcommand{\arraystretch}{1.3}
\setlength{\tabcolsep}{2pt}
\setlength{\heavyrulewidth}{0.55pt}
\setlength{\lightrulewidth}{0.35pt}
\setlength{\cmidrulewidth}{0.30pt}
\begin{tabular*}{\textwidth}{@{\extracolsep{\fill}}llcccccc@{}}
\toprule
\multirow{2}{*}{\shortstack[c]{Traffic\\Density}}
& \multicolumn{1}{c}{\multirow{2}{*}{Algorithm}}
& Success Rate
& Crash Rate
& Merging Time
& Average Velocity
& Min TTC
& TTC Violation Rate \\
&
&
(\%) $\uparrow$
& (\%) $\downarrow$
& (s) $\downarrow$
& (m/s) $\uparrow$
& (s) $\uparrow$
& (\%) $\downarrow$ \\
\midrule
\multirow{8}{*}{\shortstack[c]{Low-density\\Traffic}}
& CARGO w/o CC
& $68.00 \pm 4.00$ & $9.67 \pm 2.89$ & $10.58 \pm 0.05$
& $22.77 \pm 0.20$ & $2.77 \pm 0.19$ & $2.53 \pm 0.71$ \\
& CARGO w/o CG
& $57.67 \pm 4.16$
& $\underline{5.33 \pm 2.08}$
& $12.46 \pm 0.24$
& $21.58 \pm 0.12$ & $3.06 \pm 0.11$
& $\underline{0.71 \pm 0.32}$ \\
& CARGO w/o CR
& $59.33 \pm 1.53$ & $7.00 \pm 1.00$ & $11.96 \pm 1.43$
& $16.17 \pm 0.17$ & $2.83 \pm 0.30$ & $1.89 \pm 0.60$ \\
& CARGO w/o CR-PO
& $\underline{86.00 \pm 4.58}$
& $7.67 \pm 4.51$ & $11.09 \pm 0.20$
& $23.00 \pm 0.10$
& $\underline{3.19 \pm 0.13}$
& $1.65 \pm 0.30$ \\
& MAPPO
& $78.67 \pm 3.06$ & $10.67 \pm 3.06$
& $\underline{10.57 \pm 0.23}$
& $\underline{23.22 \pm 0.19}$
& $2.99 \pm 0.11$ & $1.67 \pm 2.05$ \\
& CARGO
& $\boldsymbol{88.00 \pm 1.00}$
& $\boldsymbol{2.00 \pm 1.00}$
& $\boldsymbol{9.47 \pm 0.21}$
& $\boldsymbol{24.85 \pm 0.08}$
& $\boldsymbol{3.70 \pm 0.22}$
& $\boldsymbol{0.48 \pm 0.05}$ \\
\midrule
\multirow{8}{*}{\shortstack[c]{Medium-density\\Traffic}}
& CARGO w/o CC
& $41.33 \pm 5.51$ & $9.67 \pm 2.89$
& $\underline{11.33 \pm 0.16}$
& $\underline{22.74 \pm 0.14}$
& $1.62 \pm 0.11$ & $3.76 \pm 0.09$ \\
& CARGO w/o CG
& $23.00 \pm 5.00$ & $16.00 \pm 4.00$ & $12.04 \pm 0.36$
& $19.86 \pm 0.09$ & $2.41 \pm 0.22$ & $2.42 \pm 0.86$ \\
& CARGO w/o CR
& $\underline{49.67 \pm 4.04}$
& $\underline{5.33 \pm 2.31}$
& $12.36 \pm 0.43$
& $16.73 \pm 0.18$
& $\underline{2.46 \pm 0.07}$
& $1.71 \pm 0.03$ \\
& CARGO w/o CR-PO
& $44.33 \pm 7.77$ & $6.67 \pm 3.21$ & $12.63 \pm 1.46$
& $18.96 \pm 0.11$ & $2.33 \pm 0.06$
& $\underline{1.43 \pm 0.13}$ \\
& MAPPO
& $48.33 \pm 1.15$ & $13.33 \pm 4.51$ & $11.87 \pm 0.01$
& $21.42 \pm 0.05$ & $2.13 \pm 0.08$ & $3.10 \pm 0.34$ \\
& CARGO
& $\boldsymbol{50.67 \pm 1.53}$
& $\boldsymbol{2.67 \pm 0.58}$
& $\boldsymbol{11.24 \pm 0.32}$
& $\boldsymbol{22.83 \pm 0.10}$
& $\boldsymbol{2.47 \pm 0.16}$
& $\boldsymbol{1.38 \pm 0.43}$ \\

\midrule

\multirow{6}{*}{\shortstack[c]{High-density\\Traffic}}
& CARGO w/o CC
& $2.33 \pm 1.15$ & $19.33 \pm 1.53$ & $14.58 \pm 0.55$
& $\underline{20.49 \pm 0.22}$
& $1.27 \pm 0.13$ & $4.68 \pm 0.97$ \\
& CARGO w/o CG
& $\underline{14.33 \pm 6.11}$
& $\underline{5.67 \pm 1.53}$
& $14.11 \pm 0.05$
& $19.19 \pm 0.18$ & $1.45 \pm 0.06$
& $\underline{2.51 \pm 0.08}$ \\
& CARGO w/o CR
& $13.00 \pm 1.73$ & $6.67 \pm 1.15$ & $13.63 \pm 1.09$
& $13.90 \pm 0.11$
& $\underline{1.78 \pm 0.11}$
& $2.73 \pm 0.19$ \\
& CARGO w/o CR-PO
& $11.00 \pm 1.00$ & $14.33 \pm 2.52$ & $12.97 \pm 0.42$
& $17.44 \pm 0.29$ & $1.61 \pm 0.06$ & $4.39 \pm 0.54$ \\
& MAPPO
& $12.33 \pm 2.89$ & $11.00 \pm 2.00$
& $\underline{12.95 \pm 0.97}$
& $18.69 \pm 0.05$ & $1.59 \pm 0.11$ & $2.92 \pm 0.18$ \\
& CARGO
& $\boldsymbol{31.00 \pm 3.61}$
& $\boldsymbol{4.67 \pm 2.89}$
& $\boldsymbol{12.74 \pm 0.81}$
& $\boldsymbol{21.97 \pm 0.14}$
& $\boldsymbol{1.89 \pm 0.03}$
& $\boldsymbol{2.37 \pm 0.11}$ \\
\bottomrule
\end{tabular*}
\begin{minipage}{\textwidth}
\scriptsize
\rmfamily
\justifying
\vspace{1mm}
\noindent
Note: CC denotes Causal Contribution,
CG denotes Causal Graph,
CR denotes Causal Responsibility,
and PO denotes Responsibility-Guided Preventive Policy Optimization. MAPPO serves as the base algorithm for CARGO.
\par
\end{minipage}

\end{table*}

We conduct an ablation study to evaluate the roles of causal interaction modeling, contribution-based incentives, and responsibility-guided preventive optimization. Four variants are considered: (i) w/o CC removes the causal contribution incentive; (ii) w/o CG replaces the inferred causal graph with a fully connected interaction graph, without identifying causal relationships; (iii) w/o CR replaces causal responsibility allocation with uniform responsibility allocation across agents while retaining preventive optimization; and (iv) w/o CR-PO removes both causal responsibility and preventive optimization while retaining the causal graph and causal contribution incentive. These comparisons examine the incremental benefit of responsibility-based allocation within the same preventive optimization framework and the role of preventive optimization under uniform allocation.

Table~\ref{tab:ablation_density} shows complementary benefits from the three components across all densities. At medium density, removing the causal graph reduces success from 50.67\% to 23.00\% and increases crashes from 2.67\% to 16.00\%. At high density, removing the contribution incentive reduces success from 31.00\% to 2.33\%. Uniform responsibility allocation reduces average velocity from 21.97 to 13.90~m/s, while removing both responsibility and preventive optimization increases crashes from 4.67\% to 14.33\%. These comparisons support contribution-based cooperation and differentiated preventive correction within the evaluated framework.

\begin{figure}[pos=!tb]
\centering
\includegraphics[width=1\linewidth]{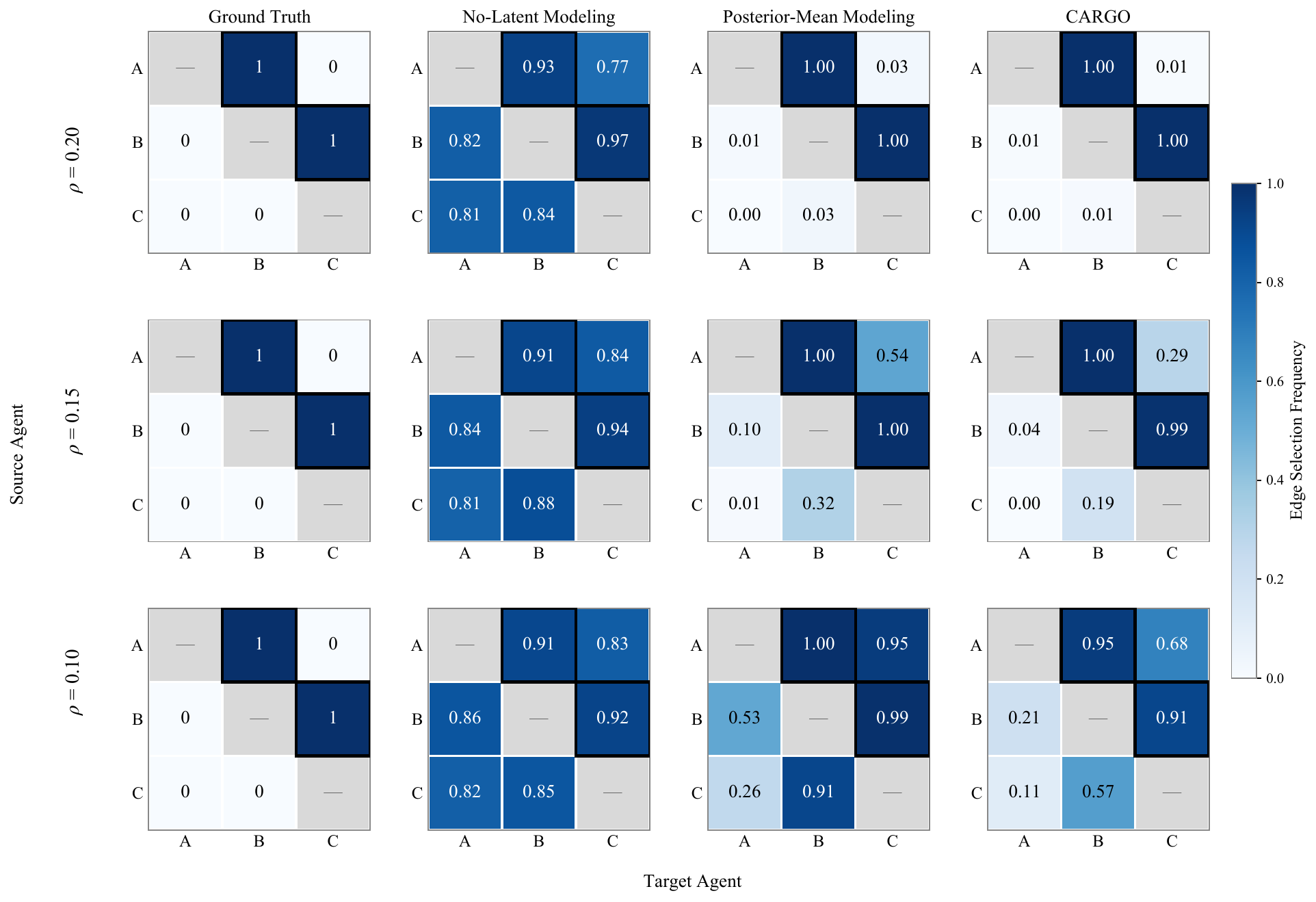}

\caption{Causal graph recovery under shared latent confounding, where the current actions of vehicles A and B influence the next actions of vehicles B and C, respectively. Rows denote source agents whose current actions exert
influence, and columns denote target agents whose
next actions are affected. Posterior-Mean Modeling learns the latent posterior but uses only its mean for graph construction, whereas No-Latent Modeling omits latent confounder representation. Entries show edge selection frequencies, ideally 1 for true edges and 0 for non-edges; gray diagonal entries are excluded. As $\rho$ decreases and direct interactions weaken, CARGO suppresses spurious edge selection more effectively than both variants while retaining most true dependencies.}
\label{fig:graph_recovery}
\end{figure}

\subsection{Analysis of Component Effectiveness}
We test graph recovery under shared latent confounding, the combination of influence and utility, responsibility-source identification, and the effect of responsibility-guided correction. Constraint-handling alternatives are compared within the same framework.

\subsubsection{Causal Graph Recovery under Latent Confounding}\label{subsubsection:Causal Graph Recovery}
To evaluate causal graph recovery under shared latent confounding, we construct a controlled experiment involving three vehicles, denoted A, B, and C, in \textit{highway-env}. The ground-truth graph contains two one-step direct dependencies: B's next action depends on A's current action, and C's next action depends on B's current action; all other non-self directed pairs are non-edges. A shared hidden preference $U$ is sampled uniformly from five actions (lane change left, maintain, lane change right, accelerate, and decelerate) and held fixed within each trajectory. At each time step, all three vehicles generate their next actions simultaneously. Vehicle A follows $U$ with probability $0.6$ and selects an action uniformly at random with probability $0.4$. Vehicles B and C copy the current action of A and B, respectively, with probability $\rho$, follow $U$ with probability $0.6$, and select an action uniformly at random with probability $0.4-\rho$.
This mechanism induces correlations between
vehicles without introducing additional direct causal edges.
We decrease $\rho$ to weaken direct interactions, thereby testing the method's ability to distinguish direct effects from spurious
correlations. Each $\rho$ setting uses 1,000 training,
200 validation, and 200 independent test trajectories. We compare full CARGO with two ablation variants.
No-Latent Modeling removes latent-variable modeling and
replaces the latent variable with a fixed zero vector.
Posterior-Mean Modeling retains latent-variable modeling
but uses only the posterior mean during inference,
without posterior sampling.

Fig.~\ref{fig:graph_recovery} presents the edge selection
frequencies. Rows and columns represent source and target agents,
respectively. Ideal recovery yields a frequency of 1 for
the two true edges and 0 for the four non-edges. No-Latent Modeling frequently selects all four noncausal edges, indicating substantial misattribution of shared influences to
direct interactions. At $\rho=0.20$, both latent-variable methods closely recover the ground-truth structure, with CARGO retaining both true edges and limiting non-edge selection frequencies to at most $0.01$. As $\rho$ decreases, the distinction
between CARGO and Posterior-Mean Modeling becomes more pronounced. At $\rho=0.15$, CARGO reduces the false-edge selection frequencies from A to C and from C to B from $0.54$ and $0.32$ to $0.29$ and $0.19$, respectively, while retaining the true edges with
frequencies of $1.00$ and $0.99$. At $\rho=0.10$, CARGO reduces the selection frequency of the false edge from A to C from $0.95$ under Posterior-Mean Modeling to $0.68$, and that from C to B from $0.91$ to $0.57$. Meanwhile, the true dependencies from A to B and from B to C are retained with selection frequencies of $0.95$ and $0.91$, respectively. These results support the effectiveness of shared latent-context modeling and posterior-aware graph construction in suppressing spurious edge selection while preserving most direct dependencies.

\subsubsection{Causal Contribution Incentive}

\begin{table*}[!t]
\centering
\caption{Validation of the causal contribution component through comparison with causal influence and team utility under medium-density traffic.}
\label{tab:cfca}
\scriptsize
\renewcommand{\arraystretch}{1.3}
\setlength{\tabcolsep}{2pt}
\setlength{\heavyrulewidth}{0.55pt}
\setlength{\lightrulewidth}{0.35pt}
\setlength{\cmidrulewidth}{0.30pt}
\begin{tabular*}{\textwidth}{@{\extracolsep{\fill}}lcccccc@{}}
\toprule
\multirow{2}{*}{Algorithm}
& Success Rate
& Crash Rate
& Merging Time
& Average Velocity
& Min TTC
& TTC Violation Rate \\
& (\%) $\uparrow$
& (\%) $\downarrow$
& (s) $\downarrow$
& (m/s) $\uparrow$
& (s) $\uparrow$
& (\%) $\downarrow$ \\
\midrule
Causal Influence
& $\underline{34.67 \pm 3.51}$
& $\underline{12.33 \pm 6.03}$
& $12.04 \pm 0.27$
& $\underline{22.15 \pm 0.26}$
& $\underline{1.27 \pm 0.11}$
& $\underline{5.33 \pm 1.05}$ \\
Team utility
& $16.67 \pm 1.15$
& $17.67 \pm 2.08$
& $\underline{11.27 \pm 0.03}$
& $19.73 \pm 0.38$
& $1.01 \pm 0.02$
& $9.10 \pm 0.05$ \\
CARGO
& $\boldsymbol{50.67 \pm 1.53}$
& $\boldsymbol{2.67 \pm 0.58}$
& $\boldsymbol{11.24 \pm 0.32}$
& $\boldsymbol{22.83 \pm 0.10}$
& $\boldsymbol{2.47 \pm 0.16}$
& $\boldsymbol{1.38 \pm 0.43}$ \\
\bottomrule
\end{tabular*}

\end{table*}
Under medium-density traffic, we compare the full contribution incentive with variants using only causal influence or team utility. Table~\ref{tab:cfca} shows success rates of 50.67\%, 34.67\%, and 16.67\%, respectively, with corresponding crash rates of 2.67\%, 12.33\%, and 17.67\%. CARGO also has the highest average velocity and minimum TTC. The combined incentive thus improves cooperation and safety relative to either signal alone in this setting.

\subsubsection{Preventive Safety Constraints}

\begin{table*}[!t]
\centering
\caption{Comparison of CARGO with MAPPO-Lagrangian~\cite{GU2023AI} and PCPO~\cite{yang2026pcpo} under low-density traffic.}
\label{tab:safety_constraints}
\scriptsize
\renewcommand{\arraystretch}{1.3}
\setlength{\tabcolsep}{2pt}
\setlength{\heavyrulewidth}{0.55pt}
\setlength{\lightrulewidth}{0.35pt}
\setlength{\cmidrulewidth}{0.30pt}
\begin{tabular*}{\textwidth}{@{\extracolsep{\fill}}lcccccc@{}}
\toprule
\multirow{2}{*}{Algorithm}
& Success Rate
& Crash Rate
& Merging Time
& Average Velocity
& Min TTC
& TTC Violation Rate \\
& (\%) $\uparrow$
& (\%) $\downarrow$
& (s) $\downarrow$
& (m/s) $\uparrow$
& (s) $\uparrow$
& (\%) $\downarrow$ \\
\midrule
CARGO + MAPPO-Lagrangian~\cite{GU2023AI}
& $30.33 \pm 2.52$
& $22.67 \pm 3.21$
& $\underline{9.81 \pm 0.42}$
& $\underline{24.04 \pm 0.07}$
& $1.32 \pm 0.10$
& $14.36 \pm 0.98$ \\
CARGO + PCPO~\cite{yang2026pcpo}
& $\underline{56.67 \pm 4.93}$
& $\underline{16.67 \pm 4.16}$
& $10.10 \pm 0.33$
& $23.80 \pm 0.13$
& $\underline{3.38 \pm 0.16}$
& $\underline{1.72 \pm 0.17}$ \\
CARGO
& $\boldsymbol{88.00 \pm 1.00}$
& $\boldsymbol{2.00 \pm 1.00}$
& $\boldsymbol{9.47 \pm 0.21}$
& $\boldsymbol{24.85 \pm 0.08}$
& $\boldsymbol{3.70 \pm 0.22}$
& $\boldsymbol{0.48 \pm 0.05}$ \\
\bottomrule
\end{tabular*}

\end{table*}

To evaluate the proposed safety constraint mechanism, we compare three constraint-handling methods within the same CARGO framework under low-density traffic, keeping the remaining components unchanged. The alternatives are MAPPO Lagrangian~\cite{GU2023AI}, which incorporates safety costs into the policy objective through adaptive Lagrange multipliers, and PCPO~\cite{yang2026pcpo}, which uses policy projection to enforce safety constraints during policy updates.

Table~\ref{tab:safety_constraints} shows that CARGO improves success from 56.67\% with PCPO to 88.00\%, while reducing crashes from 16.67\% to 2.00\%. Average velocity increases from 23.80 to 24.85~m/s and minimum TTC from 3.38 to 3.70~s. The comparison supports the proposed constraint mechanism under the evaluated low-density conditions.

\section{Discussion and Conclusion}
\label{sec:discussion}\label{sec:conclusion}
This study presents CARGO for cooperative on-ramp merging in mixed traffic. Shared-context posterior uncertainty guides interaction selection; complementary contribution and responsibility signals support cooperative incentives and targeted safety correction. Capacity-calibrated risk-reduction requirements, safety margins, and joint verification coordinate preventive policy updates. Across three highway-env traffic densities, CARGO achieves the highest success rates and lowest crash rates among the evaluated methods. Under high-density traffic, success improves by 66.04\% and crashes decrease by 56.23\% relative to the strongest baseline for each metric. Controlled experiments further support graph recovery, responsibility attribution, and differentiated correction under the tested conditions.

The reliability of these mechanisms depends on the learned response and risk models. Longer prediction horizons can accumulate model error, and additional samples reduce sampling variability without eliminating model bias. Theoretical guarantees remain conditional on local projection assumptions and valid estimation-error control; empirical safety margins do not establish unconditional safety. Future work will address the accuracy--computation trade-off in counterfactual risk estimation and scalability to larger multi-agent systems.

\bibliographystyle{cas-model2-names}
\bibliography{main.bib}
\end{document}